\pdfoutput=1

\documentclass[11pt]{article}

\usepackage{ACL2023}

\usepackage{times}
\usepackage{latexsym}
\usepackage{ulem}
\usepackage{amssymb}
\usepackage{caption}
\usepackage{subcaption}

\usepackage[T1]{fontenc}

\usepackage[utf8]{inputenc}

\usepackage{microtype}

\usepackage{inconsolata}
\usepackage{enumitem,kantlipsum}
\usepackage{enumitem}%
\usepackage{booktabs}
\usepackage{multicol}
\usepackage{multirow}
\usepackage{graphicx}
\usepackage{amsmath}
\usepackage{svg}

\newcommand\ti[1]{\textit{#1}}

\newcommand\tf[1]{\textbf{#1}}
\newcommand\ttt[1]{\texttt{#1}}

\renewcommand{\paragraph}[1]{\vspace{0.2cm}\noindent\textbf{#1}}

\definecolor{ccon}{HTML}{fee9d4}
\definecolor{cood}{HTML}{d8f0d3}
\definecolor{cid}{HTML}{dae8f5}

\definecolor{gred}{HTML}{cc0200}
\definecolor{ggreen}{HTML}{38761c}

\definecolor{c1}{cmyk}{0,0.6175,0.8848,0.1490}
\definecolor{c2}{cmyk}{0.1127,0.6690,0,0.4431}
\definecolor{c3}{cmyk}{0.3081,0,0.7209,0.3255}
\definecolor{c4}{cmyk}{0.6765,0.2017,0,0.0667}
\definecolor{c5}{cmyk}{0,0.8765,0.7099,0.3647}

\newcommand{\vanilla}{\textsc{Gold}}
\newcommand{\randoml}{\textsc{Random}}
\newcommand{\abstractl}{\textsc{Abstract}}

\newcommand{\tr}{task recognition}
\newcommand{\tl}{task learning}
\newcommand{\trabbr}{TR}
\newcommand{\tlabbr}{TL}
\newcommand{\trc}{Task recognition}
\newcommand{\tlc}{Task learning}
\newcommand{\trcc}{Task Recognition}
\newcommand{\tlcc}{Task Learning}

\newcommand{\ndatasets}{{$16$}}

\title{What In-Context Learning ``Learns'' In-Context: \\ Disentangling \trcc{} and \tlcc{}}

\author{Jane Pan \quad Tianyu Gao \quad Howard Chen \quad Danqi Chen \\
Department of Computer Science, Princeton University\\
\ttt{\{jp7224,tianyug,howardchen,danqic\}@cs.princeton.edu}\\
}

\begin{document}
\maketitle

\begin{abstract}

Large language models (LLMs) exploit in-context learning (ICL) to solve tasks with only a few demonstrations, but its mechanisms are not yet well-understood. 
Some works suggest that LLMs only recall already learned concepts from pre-training, while
others hint that ICL performs implicit learning over demonstrations.
We characterize two ways through which ICL leverages demonstrations. %
\textit{\trc{}} (\trabbr{}) captures the extent to which LLMs can 
recognize a task through demonstrations -- even without ground-truth labels~-- and apply their pre-trained priors, whereas 
\textit{\tl{}} (\tlabbr{}) is the ability to capture new input-label mappings unseen in pre-training.
Using a wide range of classification datasets and three LLM families (GPT-3, LLaMA and OPT), we design controlled experiments to disentangle the roles of \trabbr{} and \tlabbr{} in ICL.
We show that 
(1) models can achieve non-trivial performance with only \trabbr{}, and \trabbr{} does not further improve with larger models or more demonstrations; 
(2) LLMs acquire \tlabbr{} as the model scales,
and \tlabbr{}'s performance consistently improves with more demonstrations in context. 
Our findings unravel two different forces behind ICL and 
we advocate for discriminating them in future ICL research due to their distinct nature.\footnote{Our code is publicly available at \url{https://github.com/princeton-nlp/WhatICLLearns}.}

\end{abstract}

\section{Introduction}

Large language models (LLMs) have demonstrated the ability to perform in-context learning (ICL), i.e., ``learning'' to perform a task purely from examples in the context without any parameter updates~\cite{brown2020language}.
This powerful and flexible phenomenon enables LLMs to be used as general-purpose models that can perform any task with a small set of labeled examples.

However, there is still no consensus on how in-context learning works.
Some previous work hypothesizes that during pre-training, LLMs implicitly learn tasks required for downstream applications, 
and the in-context demonstrations merely provide information that allow the model to recognize which task is required~\cite{xie2022icl}. 
\citet{min2022rethinkICL} show empirical evidence of this hypothesis by demonstrating that ICL performance is insensitive to the usage of ground-truth labels.%

\newcommand{\exampletable}{   
    \resizebox{0.85\columnwidth}{!}{%
    \begin{tabular}{ll}
        \toprule
        \textbf{Setting} & \textbf{Example}\\
        \midrule
        \vanilla{} & I like the movie.\textbackslash{}n Sentiment: \textit{positive}\\
        \randoml{}& I like the movie.\textbackslash{}n Sentiment: \textit{negative}\\
        \abstractl{} & I like the movie.\textbackslash{}n Label: 1\\
        \bottomrule
    \end{tabular}%
    }}

\begin{figure}[t]
    \centering
    \includegraphics[width=0.9\columnwidth]{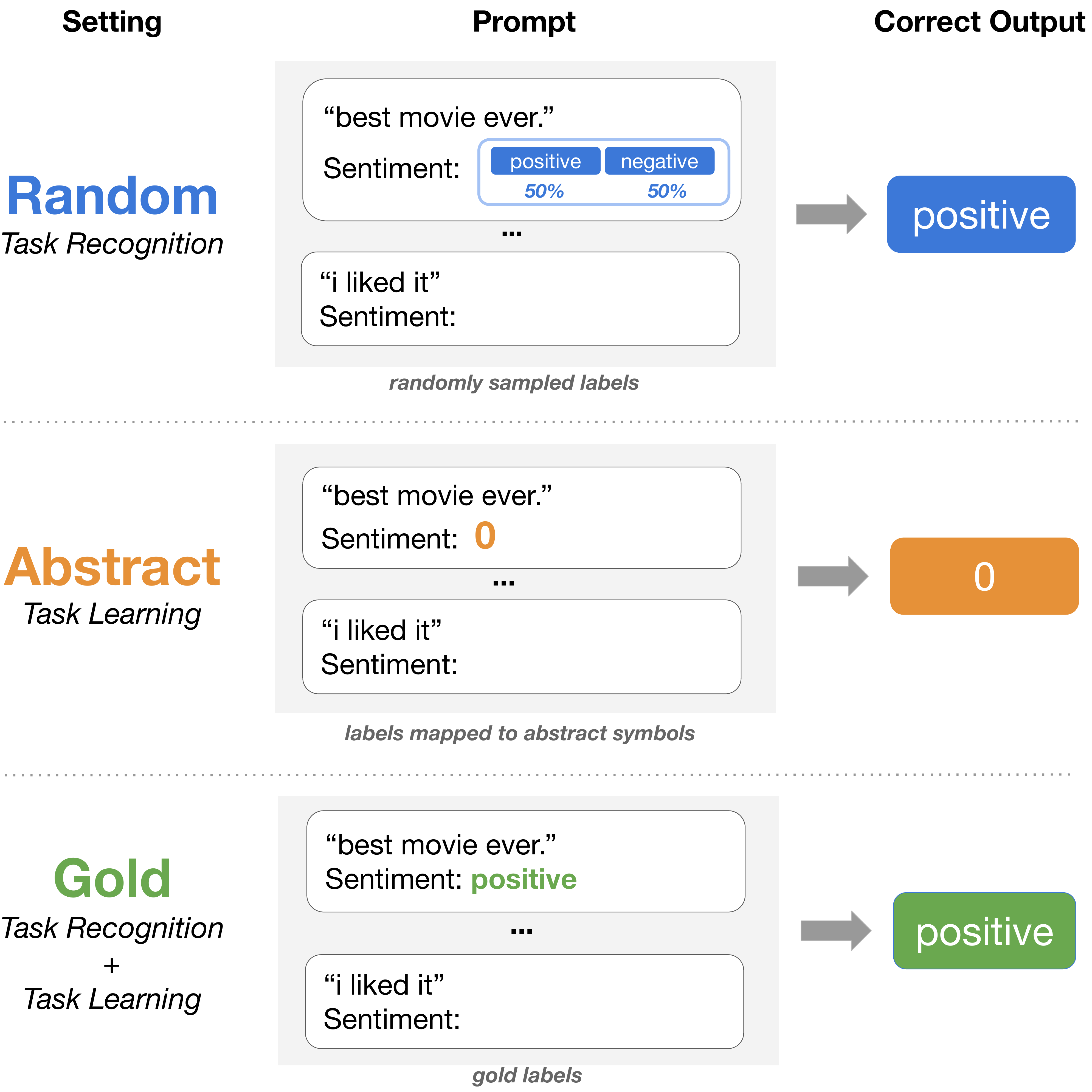}
    \caption{
        We perform experiments in three settings: \randoml{} (top), \abstractl{} (middle), and \vanilla{} (bottom). 
        Our experiments demonstrate that \textit{task recognition} (TR; shown by \randoml{}) does not scale with model sizes and number of demonstrations, while \textit{task learning} (TL; shown by \abstractl{}) does.
        }
    \vspace{-5pt}
    \label{fig:teaser}
\end{figure}

On the other hand, 
\citet{akyurek2022learning,von2022transformers} construct theories that Transformer-based models may perform implicit gradient descent to update an ``inner-model'', 
and \citet{dai2022can} demonstrate similarities between in-context learning and explicit fine-tuning through a series of metrics on real-world datasets.
Such hypotheses assume the correct input-output mappings are important and ICL actually performs implicit learning over demonstrations.

In this paper, %
we %
disentangle ICL into \textbf{\tr{}} (\trabbr{}), 
which recognizes the task from demonstrations and applies LLMs' pre-trained priors, and
\textbf{\tl{}} (\tlabbr{}), 
which learns a new input-label mapping from demonstrations.
In common ICL scenarios where ground-truth labels are provided, \trabbr{} and \tlabbr{} take effect simultaneously. We propose two settings to tease them apart:
1) \randoml{}, 
where the labels are uniformly %
sampled from the label space~\cite{min2022rethinkICL}, 
in order to restrict LLMs to only apply \trabbr{};
2) \abstractl{}, where the labels are replaced with abstract symbols 
(e.g., numbers or letters)
that never co-occurred with the inputs in pre-training.
We focus on how the two abilities in ICL evolve with two factors -- \textit{model sizes} and \textit{numbers of demonstrations}, 
which have been neglected in related literature. %

Through extensive experiments with a series of classification datasets
on GPT-3~\cite{brown2020language}, LLaMA~\cite{touvron2023llama}, and OPT~\cite{zhang2022opt},
we find: \vspace{-0.5em}
\begin{itemize}[wide,labelwidth=!,labelindent=0pt,noitemsep]

    \item The gap between \vanilla{} and \randoml{} is small 
    with smaller models,
    corroborating with  \citet{min2022rethinkICL}.
    However, with larger models and more examples,
    the gap becomes larger. %
    This suggests \trabbr{} plays a significant role in ICL, but it does not scale with increasing parameters or examples.

    \vspace{1pt}
    \item LLMs also perform \tlabbr{}, 
    which emerges with 
    larger models and more demonstrations. %
    With the largest model and more than 16 examples, 
    \abstractl{} outperforms \randoml{}, %
    pointing to a paradigm shift in in-context learning at scale.

\end{itemize}
\vspace{-0.5em}
Together, our findings provide a better way to understand ICL behaviors.\footnote{
We discuss the differences between our work and \citet{min2022rethinkICL,yoo2022deeperLook} in Section~\ref{sec:related_work}, detailing how our findings deviate and converge with existing results.}

\begin{figure*}[ht]
    \centering
    \includegraphics[width=0.92\textwidth]{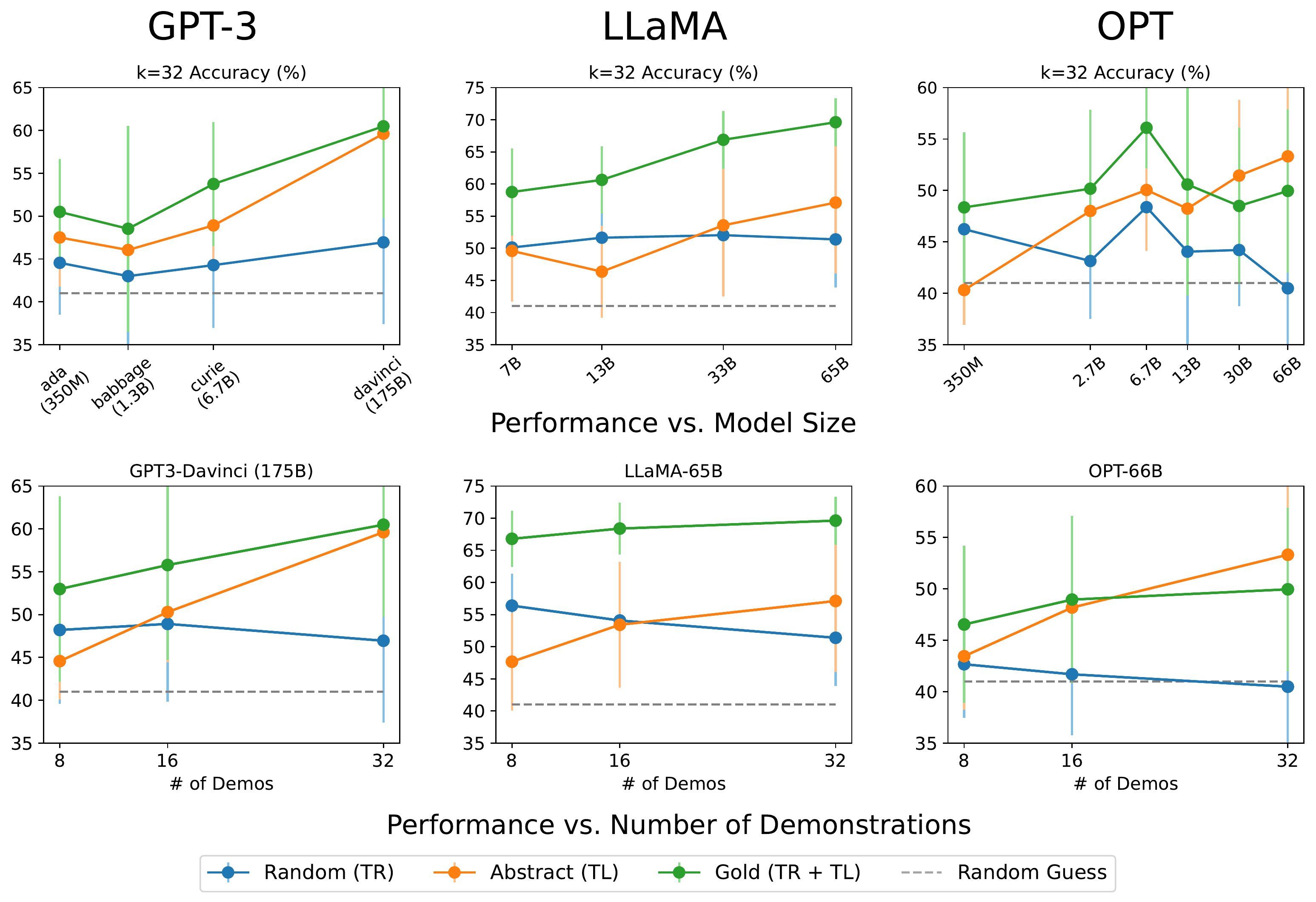}
    \caption{Averaged accuracy across \ndatasets{} datasets for GPT-3 (left), LLaMA (middle), and OPT (right).
    Top graphs plot model sizes from small to large against performance, using 32 examples.
    Variance is calculated across three prompts. Bottom graphs plot \#demonstrations against performance for \texttt{davinci}, LLaMA-65B, and OPT-66B.
    } 
    \label{fig:main_results}
\end{figure*}

\section{Task Recognition and Task Learning}\label{sec:tr_tl}

An LLM (parameterized by $\theta$) performs ICL
by conditioning on the input-label pair demonstrations 
$\mathcal{D}_\text{demo} = (x_1, y_1, x_2, y_2, \dots, x_K, y_K)$ 
and the test input $x_{\text{test}}$ to predict the label 
$y_\text{test} \sim p_\theta( y \mid \mathcal{D}_\text{demo }, x_\text{test})$, 
where the demonstrations elicit a mapping $f : \mathcal{X} \rightarrow \mathcal{Y}, x \in \mathcal{X}, y \in \mathcal{Y}$.
We delineate two ways an LLM can leverage in-context demonstrations: 
\textit{\tr{}} and \textit{\tl{}}.

\trc{} (\trabbr{}) represents models' ability to recognize the mapping $f$ purely by observing 
the input distribution $\{ x_i \}_{i=1}^K$ and the label distribution $\{ y_i\}_{i=1}^K$, without the provided $(x_i, y_i)$ pairs. 
The LLM then applies its pre-trained priors to the recognized $f$. 
Formally, when only \trabbr{} is enabled, 
\begin{align*}
    &p_\theta(y\mid x_\text{test}, \{x_i, y_i\}_{i=1}^K)  \\
    =&p_\theta(y\mid x_\text{test}, \{x_i\}_{i=1}^K, \{y_i\}_{i=1}^K), 
\end{align*}
which suggests \trabbr{} does not rely on the pair information. %
For example, an input distribution of movie reviews and a label distribution of ``The sentiment is positive/negative'' %
can be easily recognized as a sentiment classification task due to their prevalence during pre-training, and LLMs can make reasonable predictions without explicitly ``learning'' the task via ground-truth demonstrations.  
This leads to observations that the model can still perform well even when we provide wrong input-label mappings, e.g., ``The movie is great. The sentiment is \textit{negative}''~\cite{min2022rethinkICL}.
\tlc{} (\tlabbr{}), on the other hand, characterizes how the model learns a new mapping from the input-label pairs through demonstrations. 
Unlike \trabbr{}, \tlabbr{} allows models to learn novel mappings and thus correct input-label pairs will be crucial.

We posit that the two mechanisms occur under separate conditions, as recognizing an already learned task is easier than learning a new mapping.
Models are able to perform \trabbr{} at a small scale, but this ability does not drastically improve with increasing model sizes and demonstrations;
on the other hand, \tlabbr{} improves significantly when model sizes and numbers of demonstrations increase. 
To show the above phenomenon, we disentangle \trabbr{} and \tlabbr{} through \textit{label space manipulation}, including three different setups (examples in Figure~\ref{fig:teaser}):
\begin{itemize}[wide,labelwidth=!,labelindent=0pt,noitemsep]
    \item \vanilla: the standard ICL setting where we use natural prompts and gold input-label pairs. This setup reflects both \trabbr{} and \tlabbr{} abilities.
    \item \randoml: similar to \citet{min2022rethinkICL}, we use the same natural prompts as \vanilla{} and sample demonstration labels uniformly at random from the label space. This setup reflects \trabbr{} only.  
    \item \abstractl: we use minimal prompts (which provide no task information) and characters with no clear semantic meanings (e.g. numbers, letters, and random symbols) 
    as the label for each class. 
    We found that even abstract labels may have biases in pre-training, e.g., ``0'' is biased towards negative. 
    Hence, for \ti{each} prompt $x_1, y_1, \ldots, x_K, y_K$, we randomly sample a 1-1 mapping $\phi: \mathcal{Y} \rightarrow \mathcal{Y}^*$ to avoid any bias, and no task-specific information is leaked in either the prompt template or the label space. To evaluate the model's \abstractl{} performance, we measure its accuracy using $\phi(y_\text{test})$ as target labels. 
    Since these input-label mappings are never seen in pre-training, it reflects the \tlabbr{} ability. 
\end{itemize}
In the following sections, we conduct comprehensive experiments with the above three different settings under two axes -- model sizes and numbers of demonstrations -- 
and show how \trabbr{} and \tlabbr{} manifest under different conditions.

\section{Experimental Setup}\label{sec:exp}

\subsection{Datasets}
We experiment on \ndatasets{} classification datasets across 4 type of tasks: sentiment analysis, toxicity detection, natural language inference/paraphrase detection, and topic/stance classification.
All datasets and references are in Appendix~\ref{appx:datasets}.
Our dataset selection largely follows \citet{min2022rethinkICL}, but we exclude multi-choice datasets since it is difficult to apply our \abstractl{} experiments on them.

\subsection{Models}
We use three state-of-the-art LLM families: GPT-3~\cite{brown2020language}, LLaMA~\cite{touvron2023llama}, and OPT~\cite{zhang2022opt}. 
We use GPT-3 %
\texttt{ada} (350M), \texttt{babbage} (1.3B), \texttt{curie} (6.7B), and \texttt{davinci} (175B) via the OpenAI API. For OPT, we use checkpoints from the Transformers library~\cite{wolf2020transformers}, with model sizes of 350M, 2.7B, 6.7B, 13B, 30B, and 66B parameters. 
For LLaMA, we use model sizes of 7B, 13B, 33B, and 65B parameters.\footnote{For GPT-3,  we use the non-instruction legacy models for fair comparison to OPT and LLaMA models. We did not run experiments on the largest OPT-175B model due to computational constraints.}

\subsection{Task Setup}\label{sec:setup}

We adopt the sample-based evaluation protocol: 
for each test example,
we sample a different set of demonstrations from the training set. 
We manually design $3$ prompt templates for each type of classification tasks in a similar style to the prompts from \citet{min2022rethinkICL}.
We report the mean  by averaging across datasets and prompts, and 
standard variation across different prompts for each datapoint. 
For GPT-3, we sample $150$ examples for each dataset. We use fewer examples due to budget constraints, and GPT-3 presents lower variance than other model families.
For OPT and LLaMA, we sample 1,350 examples for all datasets. %

We design two kinds of prompts: \textit{natural language prompts} (Table~\ref{tab:natural_prompt}), which are similar to the manual prompts in \citet{min2022rethinkICL}, and \textit{minimal prompts} (Table~\ref{table:min_prompts}), which remove any natural language instructions for the task.
For \abstractl{}, we tested three types of label choices: \textit{numbers} ($0, \dots, N-1$, where $N$ is the number of classes), \textit{letters} ($N$ letters from A, B, C, $\dots$), and \textit{symbols} (first $N$ symbols of ``@'', ``\#'', ``\$'', ''\%'', ``*'', and ``$\wedge $''). 
For each test example, we randomly sample a new mapping between labels and abstract characters. 
We report the \textit{number} abstract labels in all the main results and compare the three forms in \S\ref{sec:result_abstract}.

\section{Results}

Figure~\ref{fig:main_results} shows our main results  with GPT-3, LLaMA, and OPT
with our $3$ settings: \vanilla{}, \randoml{}, and \abstractl{}. Below we summarize the trends of TR and TL across different conditions.

\subsection{Main Results}
\paragraph{Summary of overall trends.} 
We first verify that \vanilla{} consistently performs the best across model families and number of demonstrations, which is expected given that the \vanilla{} setting provides the model with all information. Overall, the \randoml{} curves do not increase with either model sizes or number of demonstrations, remaining largely flat; 
considering the scenario with \textit{small} model sizes and \textit{few} examples ($K=8$), 
there is an insignificant gap between \randoml{} and \vanilla{}. 
Meanwhile, the \abstractl{} curves demonstrate an increasingly steep slope 
as the model sizes and the number of demonstrations grow; with small models or small $K$, 
\abstractl{} mostly underperforms \randoml{}, 
whereas \abstractl{} with largest models and $K=32$ performs well above \randoml{} (and may even be competitive with \vanilla{}). 
We note that the OPT curves demonstrate significant variance, which we hypothesize to be a result of the models potentially being under-trained. We elaborate the takeaways on TR and TL below.

\paragraph{Task recognition is a broader capability across scales.}
For all model families,
the \randoml{} setting shows similar performance at all sizes and numbers of demonstrations.
Moreover, TR performance is significantly stronger than the random baseline, even with small models and few examples. For instance, even the smallest 350M parameter models are able to recognize the task using just $8$ examples, drawing around $10$ points of average performance lead against the random baseline for GPT-3 \texttt{ada} and $5$ points for OPT-350M. 
This shows that task recognition from in-context examples does not drastically scale with model sizes or numbers of examples.

\begin{figure}[t]
    \centering
    \includegraphics[width=1.05\columnwidth]{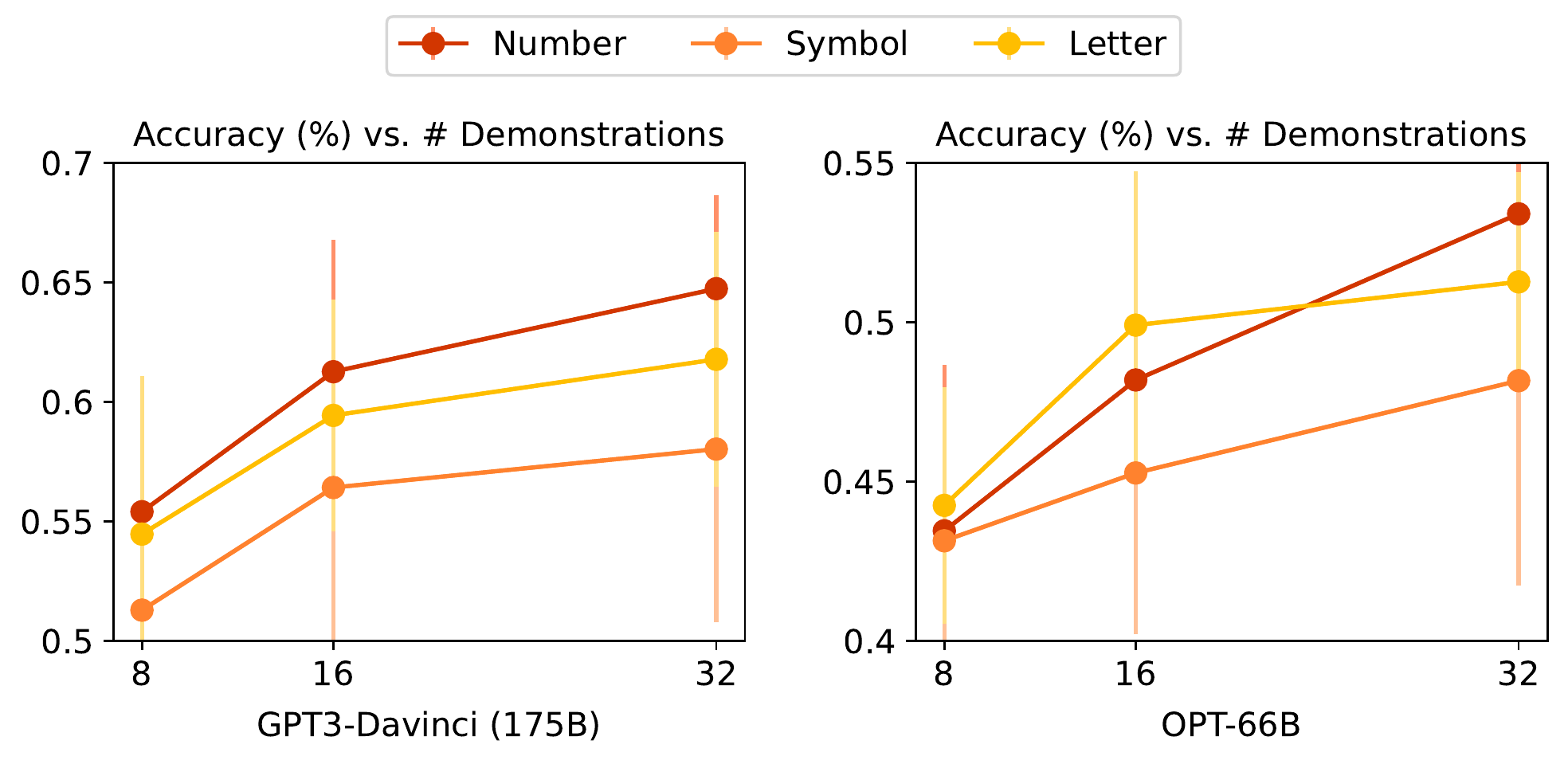}
    \caption{
    Performance of three types of \abstractl{} labels: numbers, letters, and symbols on \texttt{davinci} and OPT-66B. 
    } 
    \label{fig:abstract}
\end{figure}

\begin{figure}[t]
    \centering
    \includegraphics[width=1.05\columnwidth]{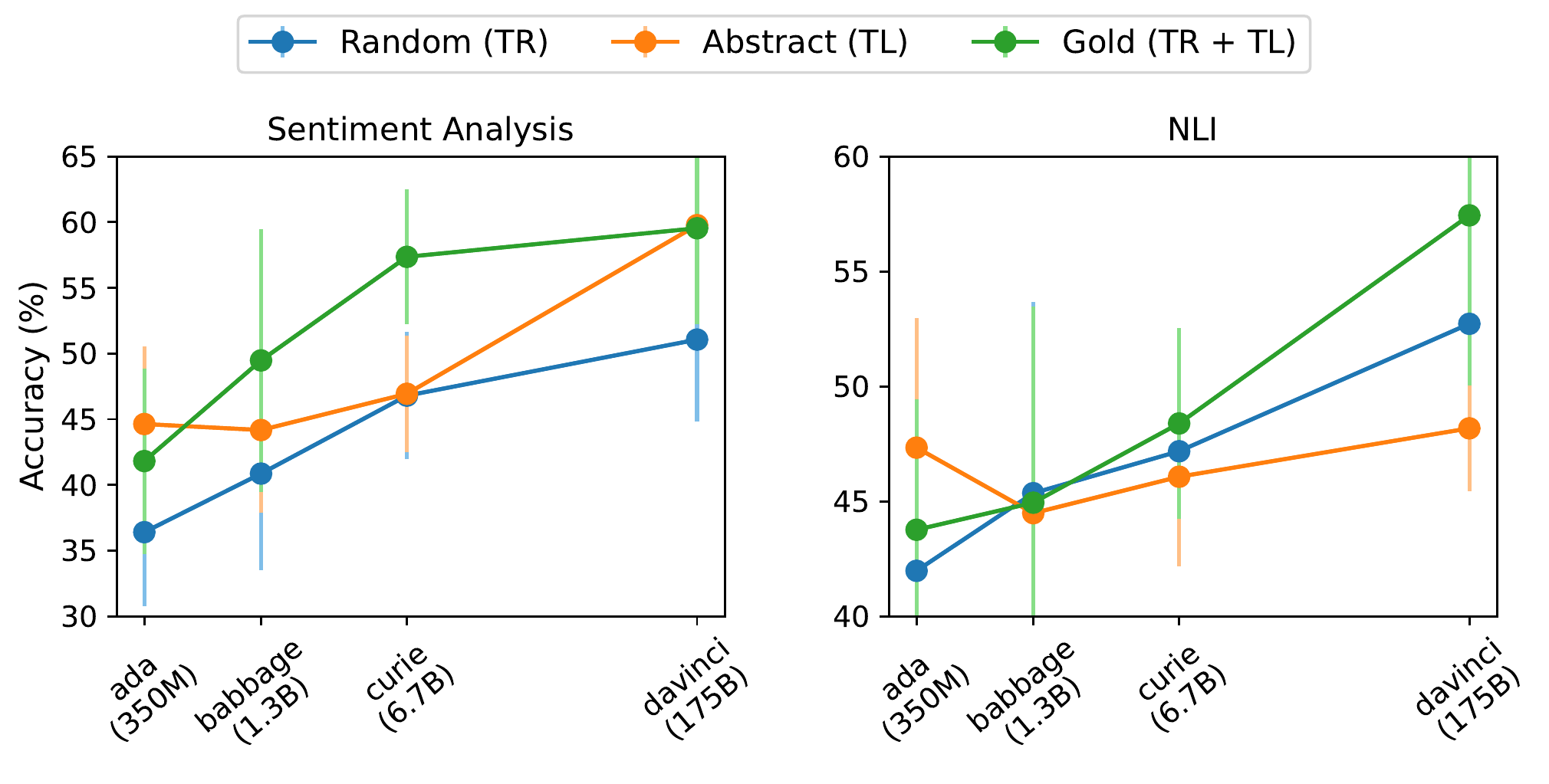}
    \caption{
    Average results of 
    sentiment analysis datasets (left) vs. natural language inference datasets (right) on GPT-3 models, with $K=32$.
    } 
    \label{fig:main_task_type}
\end{figure}

\paragraph{Task learning is enabled with scale.}
We observe that TL is dependent on model sizes:
smaller models perform roughly the same across all numbers of demonstrations (see Figure \ref{fig:demo}).
On the other hand, larger models can utilize the provided mapping information and perform TL,
as \abstractl{} (TL) performance increases drastically with larger sizes (first row of Figure~\ref{fig:main_results}).
When using a larger model, the results also improve as the number of demonstration increases (second row of Figure~\ref{fig:main_results}). 
With only $16$ examples, OPT-66B and \texttt{davinci} are able to match the performance of \vanilla{} while using a new label mapping. While LLaMA-65B's \abstractl{} is not as competitive as its \vanilla{}, the trend of improving \abstractl{} performance with larger size s or larger $K$ is clear.
This suggests that 
TL is only enabled by scales and 
further improves with more demonstrations.

\subsection{Further Analysis}
\paragraph{The trends for task learning generalize across different types of abstract labels.}
\label{sec:result_abstract}
In Figure~\ref{fig:abstract}, we show \abstractl{} results with number, letter, and symbol labels respectively. We observe that all three versions show a similar trend and coincide with our main results. Numbers and letters perform consistently better than symbols. 
This may be because letters and numbers  appear more frequently in the pre-training corpus, and  therefore make for a more "natural" label space. %

\paragraph{Task difficulty affects the trends.}
We notice that \abstractl{} scales better with sizes and examples when the task is simpler. 
In Figure~\ref{fig:main_task_type} we compare two types of tasks: sentiment analysis and natural language inference (NLI). Since NLI is more difficult, we observe that it produces a flatter \abstractl{} curve, suggesting that the model relies more on the natural prompts and pre-training priors to solve those tasks. 
We demonstrate the full task-type breakdown results in \S\ref{app:more}.

\section{Related Work}
\label{sec:related_work}

Many works have attempted to deepen empirical or theoretical understanding of ICL since its emergence in \citet{brown2020language}. For instance, \citet{xie2022icl} present a theoretical framework where latent ``concepts" parameterize each document in pre-training. They posit that all concepts have been learned in pre-training; thus, ICL is the result of implicit Bayesian inference, where the LM uses in-context demonstrations as evidence to identify the correct concept. 
\citet{min2022rethinkICL} present empirical evidence for this framework by showing that only limited information, rather than true input-label mappings, is needed to perform ICL. 

Other works investigate the impact of the pre-training corpus on ICL. \citet{chan2022data} identify properties of the pre-training distribution that enable ICL behavior, including burstiness, label multiplicity, and a long-tailed class distribution -- all of which are satisfied by natural language. \citet{razeghi2022impact} show that the frequencies of terms in the pre-training corpora is positively correlated with model performance. 

More recently, several works have explored theoretical frameworks in which ICL can be seen as implicit gradient descent, treating a
forward pass over the in-context demonstrations as an ``update” to an implicit internal model.  \citep{akyurek2022learning,von2022transformers,dai2022can}.
For mechanistic perspectives on ICL, \citet{olsson2022incontext} and \citet{bansal2022rethinking} identify induction heads (subnetworks that perform in-context pattern recognition) in small and large models, respectively.

While our conclusions align with aspects of previous studies, our work contributes novel insights on multiple axes.
\citet{min2022rethinkICL} also show that even small models can perform \trabbr{} and argue that the performance gap between \vanilla{} and \randoml{} is consistently small, but most of their experiments are on $\leq${13B} models with 16 demonstrations; we show that as model sizes scale, \vanilla{} tends to improve while \randoml{} does not. Thus, the performance deficit of \randoml{} grows as models become larger.
\citet{yoo2022deeperLook} also perform similar experiments to \randoml{} and \abstractl{}, but they do not deeply investigate the effect of model sizes or numbers of demonstrations.  Contemporary work \citet{wei2023larger} obtain similar results; additionally, they show that instruction-tuning strengthens the model's semantic priors more than it improves \tlabbr{}. However, they primarily focus on closed-source models, whereas we also conduct experiments on public models such as LLaMA and OPT. 
Collectively, our findings offer a comprehensive understanding of how ICL works across scales.

\section{Conclusion}

While previous work often studies ICL as an umbrella term, regardless of model sizes and numbers of examples,
we argue that there are two distinct characterizations of ICL -- \tr{} and \tl{} -- and demonstrate that they emerge under different conditions. 
Even small models are capable of performing \trabbr{}, but this ability does not scale. On the other hand,
\tlabbr{} is an emergent ability of large models; small models are unable to perform \tlabbr{} even when provided with more demonstrations, whereas large models can leverage more demonstations to improve their \tlabbr {} performance.
We suggest that future work on ICL should
distinguish the two phenomena and clearly state the conditions under which the experiments are conducted.

\section*{Limitations}
Though LLMs with in-context learning are capable of all kinds of NLP tasks, 
this work is limited to  classification tasks 
because they are easier to be adapted to our \randoml{} and \abstractl{} setup. 
We leave other types of NLP tasks as future work.

Another limitation of our work lies in the definition and discussion of \tl{}. 
Though we empirically show that large models are capable of acquiring a novel mapping to abstract labels like numbers or letters,  
how models ``learn'' mechanistically is still elusive. 
As suggested in previous work, LLMs may conduct implicit gradient descent over demonstrations, or they may alternatively map the patterns shown in the demonstrations back to concepts learned in pre-training. To some extent, these mechanisms could be considered an advanced form of ``task recognition''.
This work only designs experiments to better observe and disentangle \trabbr{} and \tlabbr{}, and we look forward to further studies that reveal more insights about the mechanistic inner-workings of these phenomena in ICL.

\section*{Acknowledgements}
We thank the members of the Princeton NLP group for their valuable advice, thoughts, and discussions. We also appreciate the helpful feedback given by the anonymous reviewers and the area chairs. This project was partially supported by the National Science Foundation under Award IIS-2211779, and a Sloan Fellowship.

\bibliography{main}
\bibliographystyle{acl_natbib}
\newpage
\clearpage
\appendix

\section{Datasets}\label{appx:datasets}
We use a total of \ndatasets{} datasets.
\textbf{Sentiment analysis} includes SST-2~\cite{socher2013recursive}, financial\_phrasebank~\cite{Malo2014GoodDO}, emotion~\cite{saravia2018carer}, and poem\_sentiment~\cite{sheng2020investigating}
\textbf{Topic/stance classification} includes TREC~\cite{voorhees2000building_trec}, tweet\_eval\_atheist, and tweet\_eval\_feminist \cite{mohammad2018semeval, basile2019semeval}.
\textbf{Toxicity detection} includes
tweet\_eval\_hate, ethos\_race, ethos\_gender, ethos\_national\_origin, and ethos\_religion~\cite{mollas2020ethos}
\textbf{Natural language inference/paraphrase detection} includes SICK~\cite{marelli2014sick}, SNLI~\cite{bowman2015large_snli}, WNLI~\cite{levesque2012winograd_wnli}, and MRPC~\cite{dolan2005automatically_mrpc}.

We sample from the training set to construct the prompts; following \citet{min2022rethinkICL}, we use the development set for evaluation, using sampled  $\max(1350, \text{dataset\_size})$ examples.

\section{Prompt Templates}\label{appx:prompt_templates}
For each task category (e.g. sentiment classification, topic detection), we manually design three natural language templates. Depending on exact specifications for the dataset, templates may be adjusted to better reflect the task (e.g. "Is this atheist?" for tweet\_eval\_atheist). We apply these templates to the natural language label sets (\vanilla{} and \randoml). 
All  prompts are presented in Table~\ref{tab:natural_prompt}.

We also design two task-agnostic variations on three minimal templates for \abstractl{}: one for single-sentence tasks and one for multi-sequence tasks (e.g. NLI tasks).
We use these minimal templates on the abstract language label sets in order to prevent the model from being exposed to any information regarding the task from the prompt design. All minimal templates are presented in Table~\ref{table:min_prompts}

All prompts are designed to be answered with single-token responses (e.g. "Yes/No", "True/False", "positive/negative/neutral", "0/1/2", "A/B/C") so that we can directly check models' last token prediction results instead of applying decoding methods.

\begin{table*}[h]
\small
\centering
\begin{tabular}{|p{2cm}|p{2cm}|p{10cm}|} 
\hline
Type               & Template \# & Example                                                                                                                                                                                                                                                                                                                              \\ 
\hline
                   & 1           & \textcolor{blue}{\textless{}s\textgreater{} }\textcolor[rgb]{0.502,0.502,0.502}{\newline}\textcolor{blue}{~}The sentiment is \textcolor{red}{\textless{}}\textit{\textcolor{red}{positive/negative}}\textcolor{red}{\textit{\textgreater{}}}                                                                                \\ 
\cline{2-3}
\begin{tabular}[c]{@{}l@{}}Sentiment\\Analysis\end{tabular} & 2           & \textcolor{blue}{\textless{}\textit{s}\textgreater{}~}\textcolor[rgb]{0.502,0.502,0.502}{\newline}~Sentiment:~\textcolor{red}{\textless{}}\textit{\textcolor{red}{positive/negative}}\textcolor{red}{\textit{\textgreater{}}}                                                                                               \\ 
\cline{2-3}
                   & 3           & \textcolor{blue}{\textless{}\textit{s}\textgreater{}~}\textcolor[rgb]{0.502,0.502,0.502}{\newline}\textcolor{blue}{~}The sentiment of the text is~\textcolor{red}{\textless{}\textit{positive/negative\textgreater{}}}                                                                                                      \\ 
\hline
                   & 1           & \textcolor{blue}{\textless{}\textit{s}\textgreater{}~}\textcolor[rgb]{0.502,0.502,0.502}{\newline}\textcolor{blue}{~}Is this hate speech?~\textcolor{red}{\textless{}\textit{Yes/No\textgreater{}}}                                                                                                                         \\ 
\cline{2-3}
Hate Speech        & 2           & \textcolor{blue}{\textless{}\textit{s}\textgreater{}~}\textcolor[rgb]{0.502,0.502,0.502}{\newline}\textcolor{blue}{~}Is the sentence hateful?~\textcolor{red}{\textless{}\textit{Yes/No\textgreater{}}}                                                                                                                     \\ 
\cline{2-3}
                   & 3           & \textcolor{blue}{\textless{}\textit{s}\textgreater{}~}\textcolor[rgb]{0.502,0.502,0.502}{\newline}\textcolor{blue}{~}The sentence contains hate speech. True or False?\textcolor{blue}{~}\textcolor[rgb]{0.502,0.502,0.502}{\newline~}The answer is~\textcolor{red}{\textless{}\textit{True/False\textgreater{}}}  \\ 
\hline
                   & 1           & \textcolor{blue}{\textless{}\textit{s}\textgreater{}~}\textcolor[rgb]{0.502,0.502,0.502}{\newline}\textcolor{blue}{~}The stance is feminist. True or False?~\textcolor[rgb]{0.502,0.502,0.502}{\newline}\textcolor{blue}{~}The answer is~\textcolor{red}{\textless{}\textit{True/False\textgreater{}}}             \\ 
\cline{2-3}
\begin{tabular}[c]{@{}l@{}}Stance\\Detection\end{tabular}   & 2           & \textcolor{blue}{\textless{}\textit{s}\textgreater{}~}\textcolor[rgb]{0.502,0.502,0.502}{\newline}\textcolor{blue}{~}Does the sentence express a feminist view?~\textcolor{red}{\textless{}\textit{Yes/No\textgreater{}}}                                                                                                   \\ 
\cline{2-3}
     & 3           & \textcolor{blue}{\textless{}\textit{s}\textgreater{}~}\textcolor[rgb]{0.502,0.502,0.502}{\newline}\textcolor{blue}{~}Is the stance feminist?~\textcolor{red}{\textless{}\textit{Yes/No\textgreater{}}}                                                                                                                      \\ 
\hline
                   & 1           & \textcolor{blue}{\textless{}\textit{s}\textgreater{}~}\textcolor[rgb]{0.502,0.502,0.502}{\newline}\textcolor{blue}{~}The topic is~\textcolor{red}{\textless{}\textit{label}}\textcolor{red}{\textit{\textgreater{}}}                                                                                                        \\ 
\cline{2-3}
\begin{tabular}[c]{@{}l@{}}Topic\\Detection\end{tabular}    & 2           & \textcolor{blue}{\textless{}\textit{s}\textgreater{}~}\textcolor[rgb]{0.502,0.502,0.502}{\newline}\textcolor{blue}{~}The sentence is about~\textcolor{red}{\textless{}\textit{label}}\textcolor{red}{\textit{\textgreater{}}}                                                                                               \\ 
\cline{2-3}
                   & 3           & \textcolor{blue}{\textless{}\textit{s}\textgreater{}~}\textcolor[rgb]{0.502,0.502,0.502}{\newline}\textcolor{blue}{~}Sentence topic:~\textcolor{red}{\textless{}\textit{label}}\textcolor{red}{\textit{\textgreater{}}}                                                                                                     \\
\hline

\end{tabular}
\caption{
Natural prompts used as input in \vanilla{} and \randoml{} settings for single-sentence datasets. \textcolor{blue}{\textless{}\textit{s}\textgreater{}} denotes the input sequence; labels are illustrated in \textcolor{red}{red}.
}
\label{tab:natural_prompt}
\end{table*}

\begin{table*}[h]
\centering
\small
\begin{tabular}{|p{2cm}|p{1.9cm}|p{10cm}|} 
\hline
\textbf{Type}                          & \textbf{Temp. \#}        & \textbf{Example}                                                                                                        \\ 
\hline
\multirow{12}{*}{Entailment}           & \multirow{4}{*}{1} & \textcolor{blue}{\textless{}\textit{s1}\textgreater{}}                                                                  \\
                                       &                    & The question is: \textcolor{blue}{\textless{}\textit{s2}\textgreater{}}?                                                \\
                                       &                    & True or False?                                                                                                          \\
                                       &                    & The answer is~\textcolor{red}{\textless{}\textit{True/False}\textgreater{}}                                             \\ 
\cline{2-3}
                                       & \multirow{4}{*}{2} & Hypothesis:~\textcolor{blue}{\textless{}\textit{s1}\textgreater{}}                                                      \\
                                       &                    & Premise:~\textcolor{blue}{\textless{}\textit{s2}\textgreater{}}?                                                        \\
                                       &                    & Do the sentences show entailment?                                                                                       \\
                                       &                    & \textcolor{red}{\textless{}}\textit{\textcolor{red}{Yes/No}}\textcolor{red}{\textgreater{}}                             \\ 
\cline{2-3}
                                       & \multirow{4}{*}{3} & The hypothesis is:~\textcolor{blue}{\textless{}\textit{s1}\textgreater{}}                                               \\
                                       &                    & The premise is:~\textcolor{blue}{\textless{}\textit{s2}\textgreater{}}?                                                 \\
                                       &                    & Is this entailment?                                                                                                     \\
                                       &                    & \textcolor{red}{\textless{}}\textit{\textcolor{red}{Yes/No}}\textcolor{red}{\textgreater{}}                             \\ 
\hline
\multirow{12}{*}{NLI}                  & \multirow{4}{*}{1} & \textcolor{blue}{\textless{}}\textit{s1}\textcolor{blue}{\textgreater{}}                                                \\
                                       &                    & The question is:~\textcolor{blue}{\textless{}\textit{s2}\textgreater{}}                                                 \\
                                       &                    & True, False, or Unknown?                                                                                                \\
                                       &                    & The answer is~\textcolor{red}{\textless{}}\textit{\textcolor{red}{True/False/Unknown}}\textcolor{red}{\textgreater{}}   \\ 
\cline{2-3}
                                       & \multirow{4}{*}{2} & Hypothesis:~\textcolor{blue}{\textless{}\textit{s1}\textgreater{}}                                                      \\
                                       &                    & Premise:~\textcolor{blue}{\textless{}\textit{s2}\textgreater{}}?                                                        \\
                                       &                    & Given the premise, is the hypothesis true? Yes, No, or Unknown?                                                         \\
                                       &                    & The answer is:~\textcolor{red}{\textless{}}\textit{\textcolor{red}{Yes/No/Unknown}}\textcolor{red}{\textgreater{}}      \\ 
\cline{2-3}
                                       & \multirow{4}{*}{3} & The hypothesis is:~\textcolor{blue}{\textless{}\textit{s1}\textgreater{}}                                               \\
                                       &                    & The premise is:~\textcolor{blue}{\textless{}\textit{s2}\textgreater{}}?                                                 \\
                                       &                    & According to the premise, the hypothesis is true. True, False, or Unknown?                                              \\
                                       &                    & The answer is:~\textcolor{red}{\textless{}}\textit{\textcolor{red}{True/False/Unknown}}\textcolor{red}{\textgreater{}}  \\ 
\hline
\multirow{12}{*}{\begin{tabular}[c]{@{}l@{}}Paraphrase\\Detection\end{tabular}} & \multirow{4}{*}{1} & \textcolor{blue}{\textless{}}\textit{\textcolor{blue}{s1}}\textcolor{blue}{\textgreater{}}                              \\
                                       &                    & The question is:~\textcolor{blue}{\textless{}\textit{s2}\textgreater{}}                                                 \\
                                       &                    & True or False?                                                                                                          \\
                                       &                    & The answer is:~\textcolor{red}{\textless{}}\textit{\textcolor{red}{True/False/\textgreater{}}}                          \\ 
\cline{2-3}
                                       & \multirow{4}{*}{2} & Sentence 1:~\textcolor{blue}{\textless{}}\textit{\textcolor{blue}{s1}}\textcolor{blue}{\textgreater{}}                  \\
                                       &                    & Sentence 2:~\textcolor{blue}{\textless{}}\textit{\textcolor{blue}{s2}}\textcolor{blue}{\textgreater{}}                  \\
                                       &                    & These sentences are paraphrases. True or False?                                                                         \\
                                       &                    & The answer is:~\textcolor{red}{\textless{}}\textit{\textcolor{red}{True/False/\textgreater{}}}                          \\ 
\cline{2-3}
                                       & \multirow{4}{*}{3} & Text:~\textcolor{blue}{\textless{}}\textit{\textcolor{blue}{s1}}\textcolor{blue}{\textgreater{}}                        \\
                                       &                    & Consider this sentence:~\textcolor{blue}{\textless{}}\textit{\textcolor{blue}{s2}}\textcolor{blue}{\textgreater{}}      \\
                                       &                    & Does it paraphrase the text?                                                                                            \\
                                       &                    & \textcolor{red}{\textless{}}\textit{\textcolor{red}{Yes/No\textgreater{}}}                                              \\
\hline
\end{tabular}
\caption{
Natural prompts used as input in \vanilla{} and \randoml{} settings for multi-sentence datasets. \textcolor{blue}{\textless{}\textit{s1}\textgreater} and  \textcolor{blue}{\textless{}\textit{s2}\textgreater} denote the input sequences; labels are illustrated in \textcolor{red}{red}. %
}
\label{tab:nli_prompt}
\end{table*}

\begin{table*}[t]
\centering
\small
\begin{tabular}{|p{2cm}|p{2.4cm}|p{8cm}|} 
\hline

\textbf{Type}                                                               & \textbf{Template \# }         & \textbf{Example}                                                                                                                                                                                                                                                                                                                                                              \\ 
\hline
                                                                   & \hspace{1mm}{\newline~}\text{1}                    & \textcolor{blue}{\textless{}\textit{sentence}\textgreater{}~}\textcolor[rgb]{0.502,0.502,0.502}{\newline~}\textcolor{red}{\textless{}\textit{label}\textgreater{}}                                                                                                                                                                                          \\ 
\cline{2-3}
\begin{tabular}[c]{@{}l@{}}Minimal\\(single \\ sentence)\end{tabular} & 2                    & \textcolor{blue}{\textless{}\textit{sentence}\textgreater{}~}\textcolor[rgb]{0.502,0.502,0.502}{\newline~}Label: \textcolor{red}{\textless{}\textit{label}\textgreater{}}                                                                                                                                                                                   \\ 
\cline{2-3}
                                                                   & \hspace{1mm}{\newline~}\text{3}                   & Sentence: \textcolor{blue}{\textit{sentence}\textgreater{}~}\textcolor[rgb]{0.502,0.502,0.502}{\newline~}Label:~\textcolor{red}{\textless{}\textit{label}\textgreater{}}                                                                                                                                                                         \\ 
\hline
                                                                   & \hspace{1mm}{\newline~}\text{1}                       & \textcolor{blue}{\textless{}\textit{sentence1}\textgreater{}~}{[}SEP]\textcolor[rgb]{0.808,0.569,0.471}{~}\textcolor{blue}{\textless{}\textit{sentence2}\textgreater{}~}\textcolor[rgb]{0.502,0.502,0.502}{\newline~}\textcolor{red}{\textless{}\textit{label}\textgreater{}}\textcolor{blue}{}                                                             \\ 
\cline{2-3}
\begin{tabular}[c]{@{}l@{}}Minimal\\(multiple \\ sentences)\end{tabular}  & 2                    & \textcolor{blue}{\textless{}\textit{sentence1}\textgreater{}~}{[}SEP]\textcolor[rgb]{0.808,0.569,0.471}{~}\textcolor{blue}{\textless{}\textit{sentence2}\textgreater{}~}\textcolor[rgb]{0.502,0.502,0.502}{\newline~}Label:~\textcolor{red}{\textless{}\textit{label}\textgreater{}}                                                                        \\ 
\cline{2-3}
                                                                   & \hspace{1mm}{\newline~}\text{3}                       & Sentence 1:~\textcolor{blue}{\textless{}\textit{sentence1}\textgreater{}~}\textcolor[rgb]{0.502,0.502,0.502}{\newline~}Sentence 2:\textcolor[rgb]{0.808,0.569,0.471}{~}\textcolor{blue}{\textless{}\textit{sentence2}\textgreater{}~}\textcolor[rgb]{0.502,0.502,0.502}{\newline~}Label:~\textcolor{red}{\textless{}\textit{label}\textgreater{}}  \\ 
\hline
\multicolumn{1}{l}{}                                               & \multicolumn{1}{l}{}                                                                                                                                                                                                                                                                                                                                                                    
\end{tabular}
\caption{Minimal prompts used for \abstractl{}.} 
\label{table:min_prompts}
\end{table*}

\section{More Results}
\label{app:more}
\begin{figure*}[ht]
    \centering
    \includegraphics[width=0.92\textwidth]{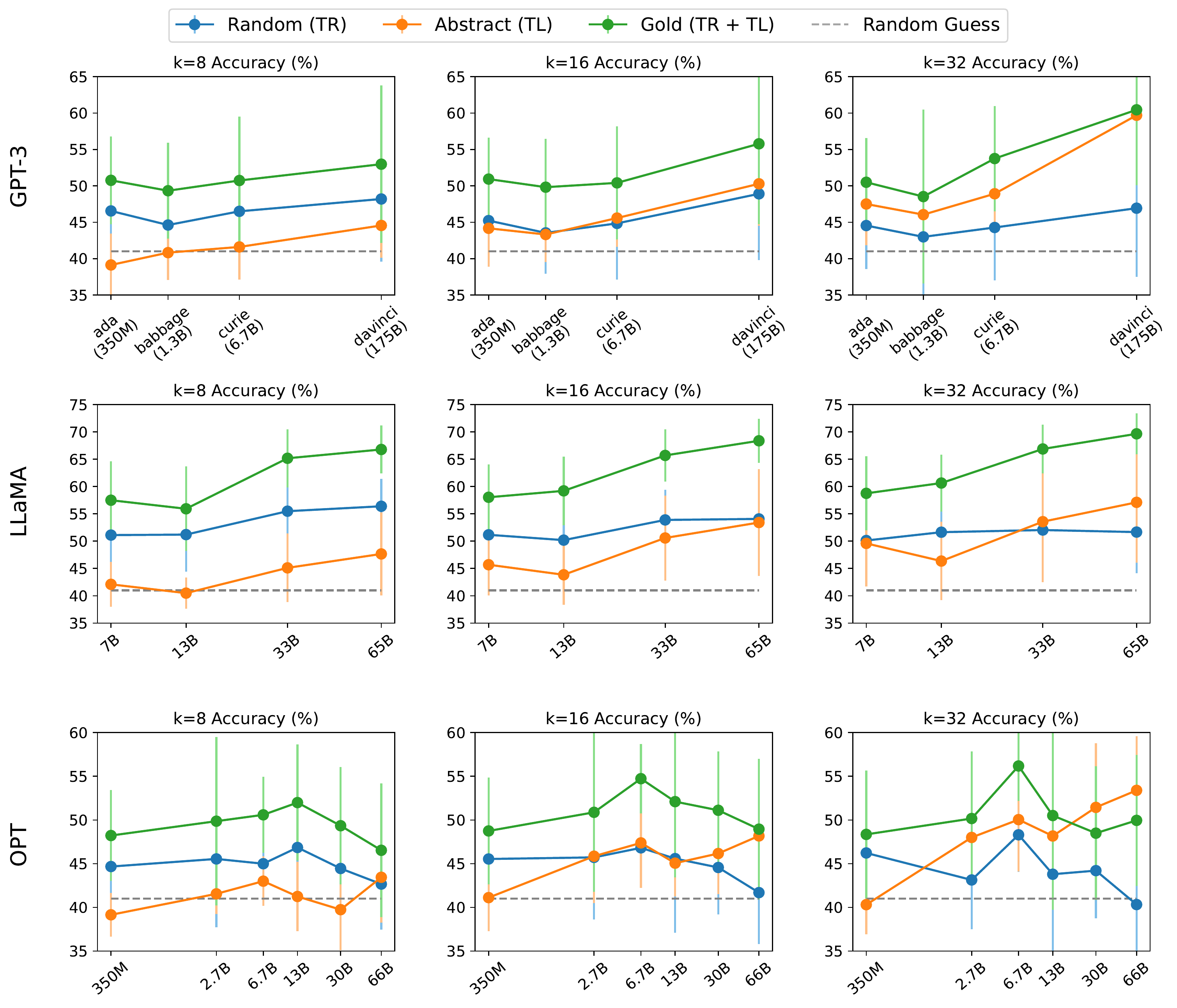}
    \caption{Averaged accuracy across \ndatasets{} datasets for GPT-3 (top), LLaMA (middle), and  OPT (bottom). X-axis shows
    model sizes from small to large. We run experiments with 8 (left), 16 (middle), and 32 (right) demonstrations respectively. Variance is calculated across three prompts.
    } 
    \label{fig:old_main_results}
\end{figure*}

We demonstrate average model performance with respect to number of parameters in Figure~\ref{fig:old_main_results}. 
It is clear that small models struggle to perform \abstractl{}, regardless of how many examples, whereas the largest models (especially GPT-3 Davinci and OPT-66B) are able to perform \abstractl{}. Additionally, their performance improves even more when more demonstrations are provided. 

\begin{figure*}[ht]
    \centering
    \includegraphics[width=0.92\textwidth]{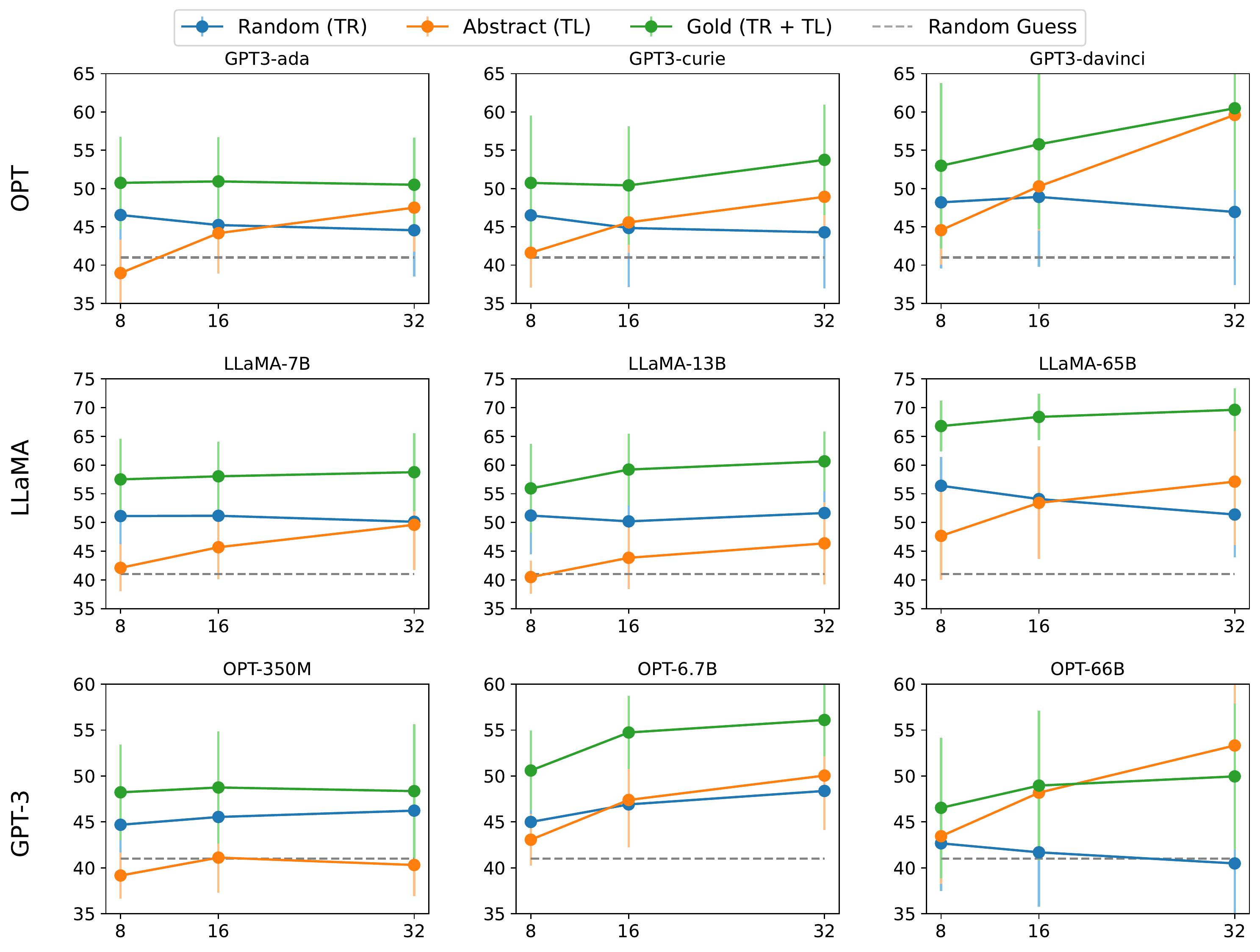}
    \caption{Averaged accuracy across \ndatasets{} datasets for GPT-3 (top), LLaMA (middle), and  OPT (bottom). x-axis shows number of demonstrations in the prompt. For each model, we run experiments for \randoml{} (left), \abstractl (middle), and \vanilla{} (right) demonstrations. Variance is calculated across three templates. 
    } 
    \label{fig:demo}
\end{figure*}

We demonstrate average model performance with respect to numbers of demonstrations in Figure~\ref{fig:demo}. 
We can see a clear trend that \randoml{} (TR) does not change much but \abstractl{} improves drastically with more examples, especially for GPT-3 Davinci and OPT-66B.

\begin{figure*}[ht]
    \centering
    \includegraphics[width=0.92\textwidth]{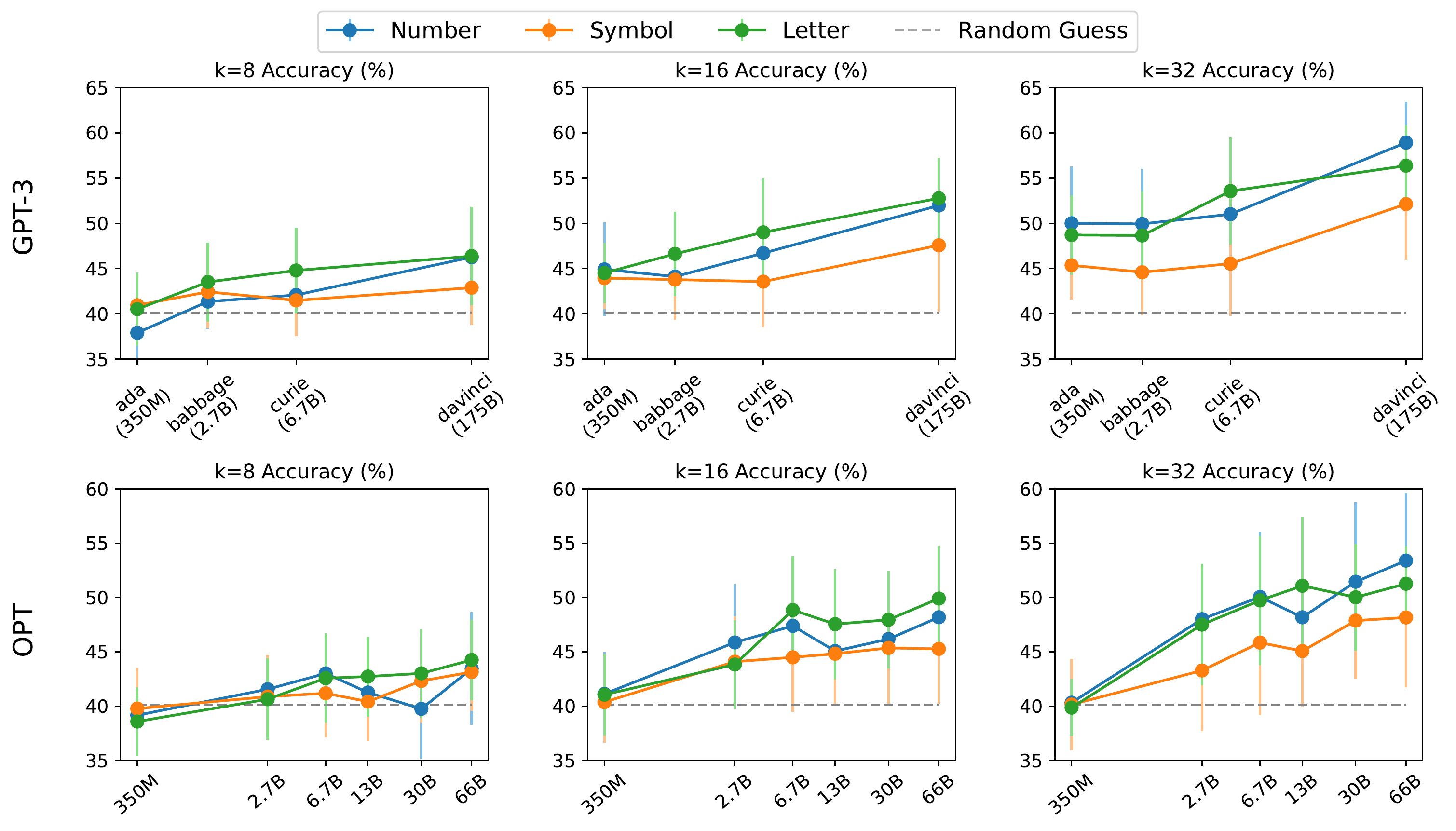}
    \caption{Performance of three types of \abstractl{} labels: numbers, letters, and symbols on OPT and GPT-3 models.
    } 
    \label{fig:all_abstract}
\end{figure*}

Figure~\ref{fig:all_abstract} shows all the \abstractl{} results and demonstrates a similar trend to what \S\ref{sec:result_abstract} describes. 

\begin{figure*}[ht]
    \centering
    \includegraphics[width=0.92\textwidth]{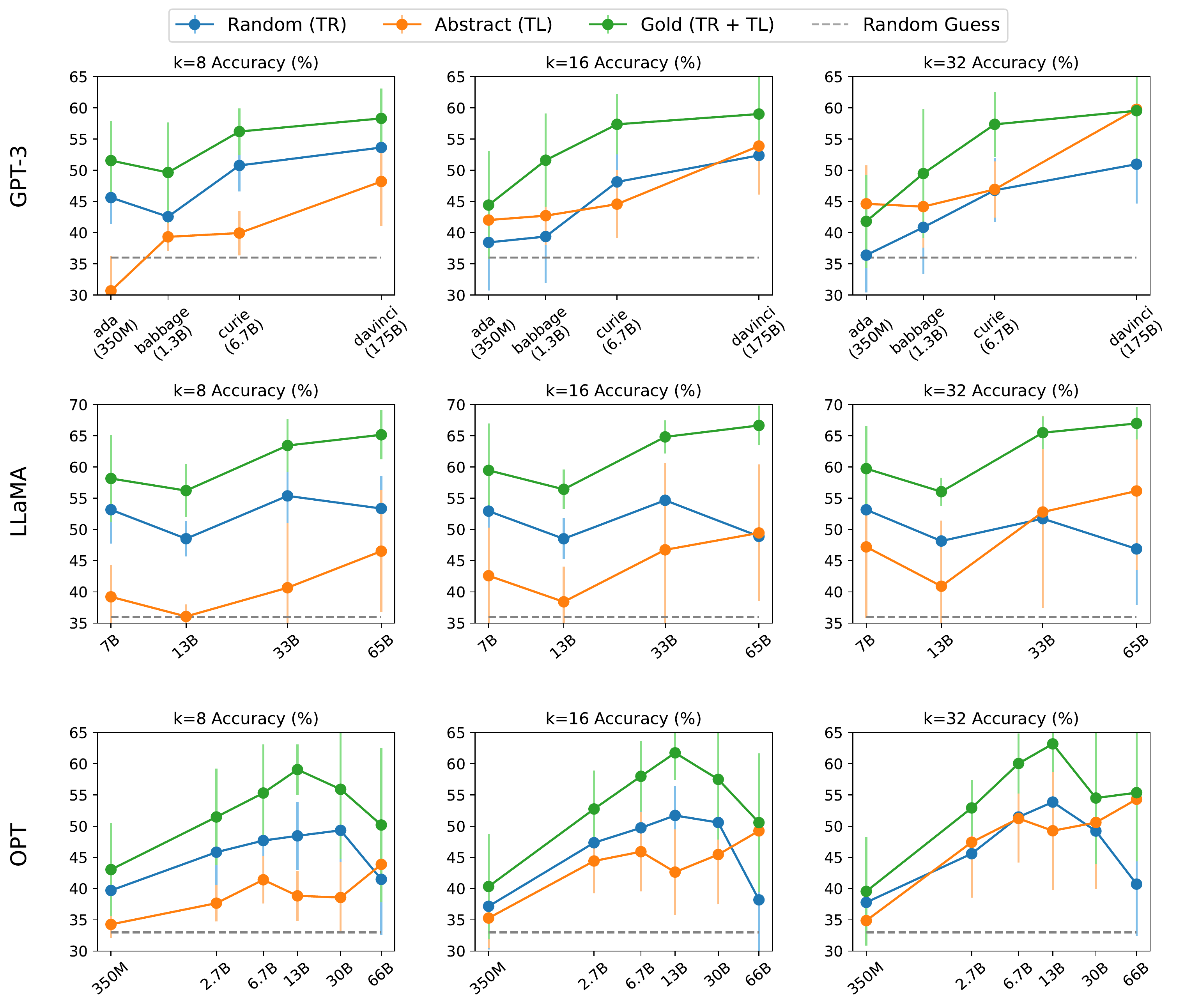}
    \caption{Average performance of \textbf{sentiment analysis} datasets.
    } 
    \label{fig:sentiment_analysis}
\end{figure*}

\begin{figure*}[ht]
    \centering
    \includegraphics[width=0.92\textwidth]{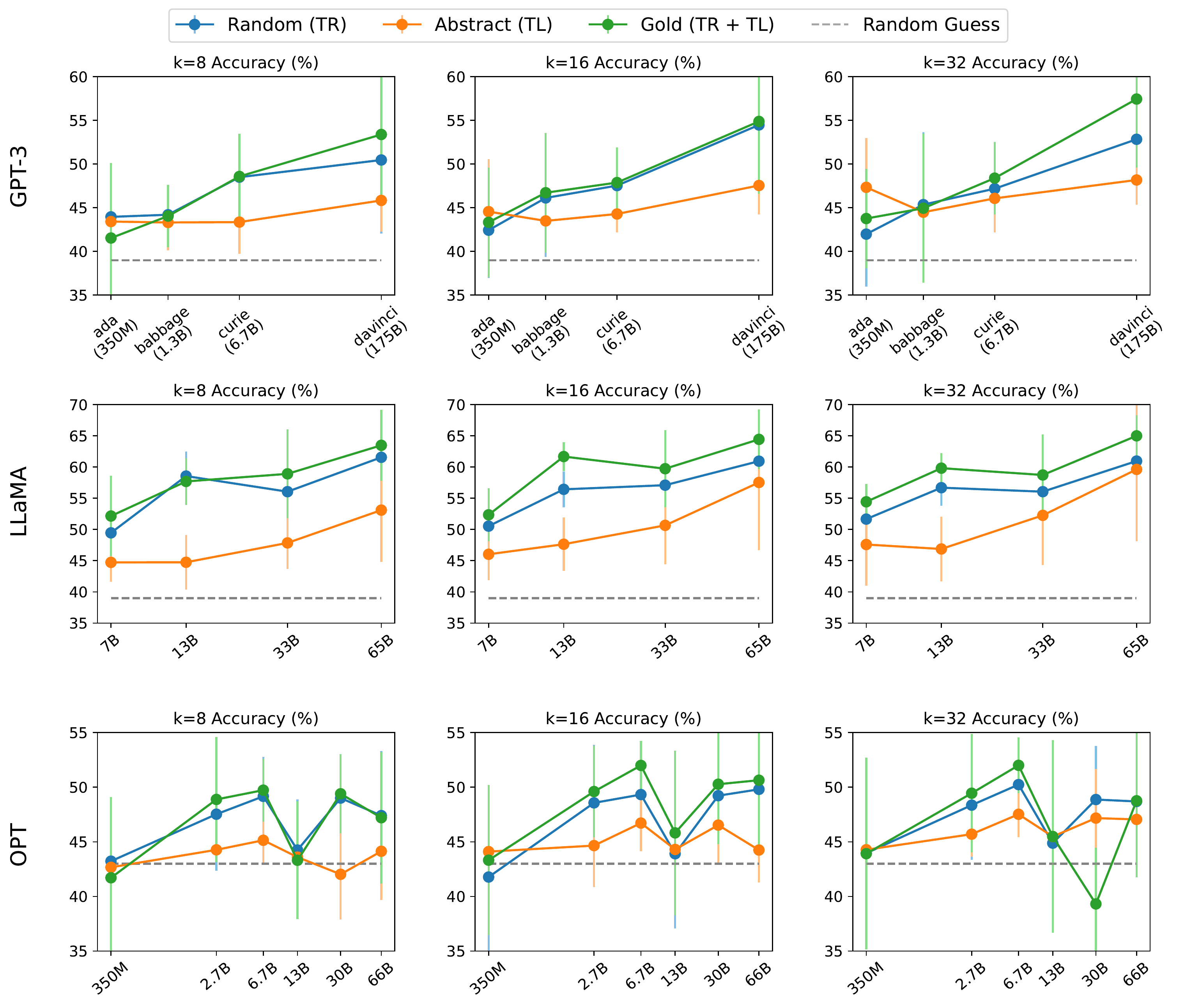}
    \caption{
    Average performance of \textbf{natural language inference/paraphrase detection} datasets. %
    } 
    \label{fig:NLI}
\end{figure*}

\begin{figure*}[ht]
    \centering
    \includegraphics[width=0.92\textwidth]{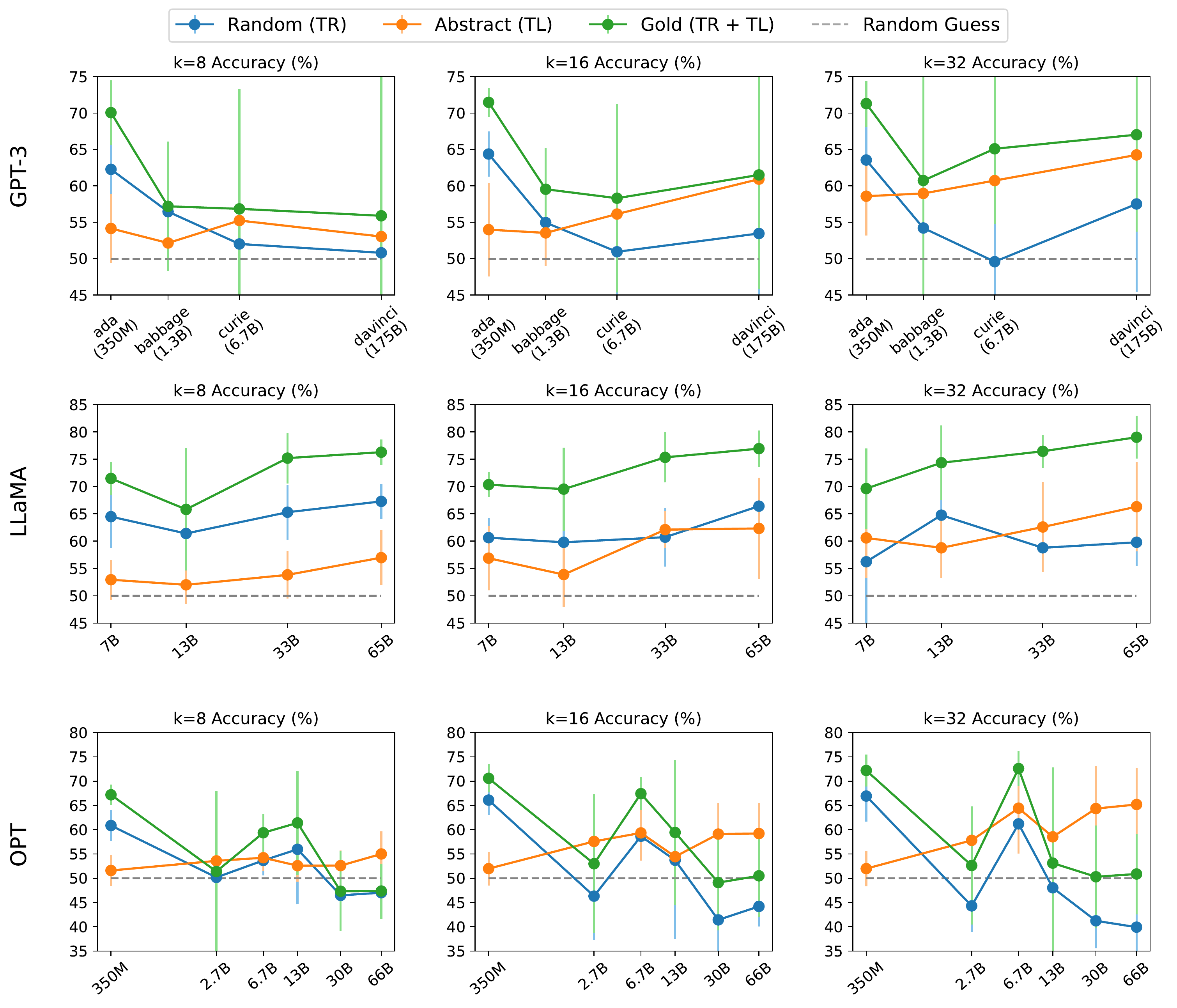}
    \caption{Average performance of \textbf{toxicity detection} datasets. %
    } 
    \label{fig:toxicity_detection}
\end{figure*}

\begin{figure*}[ht]
    \centering
    \includegraphics[width=0.92\textwidth]{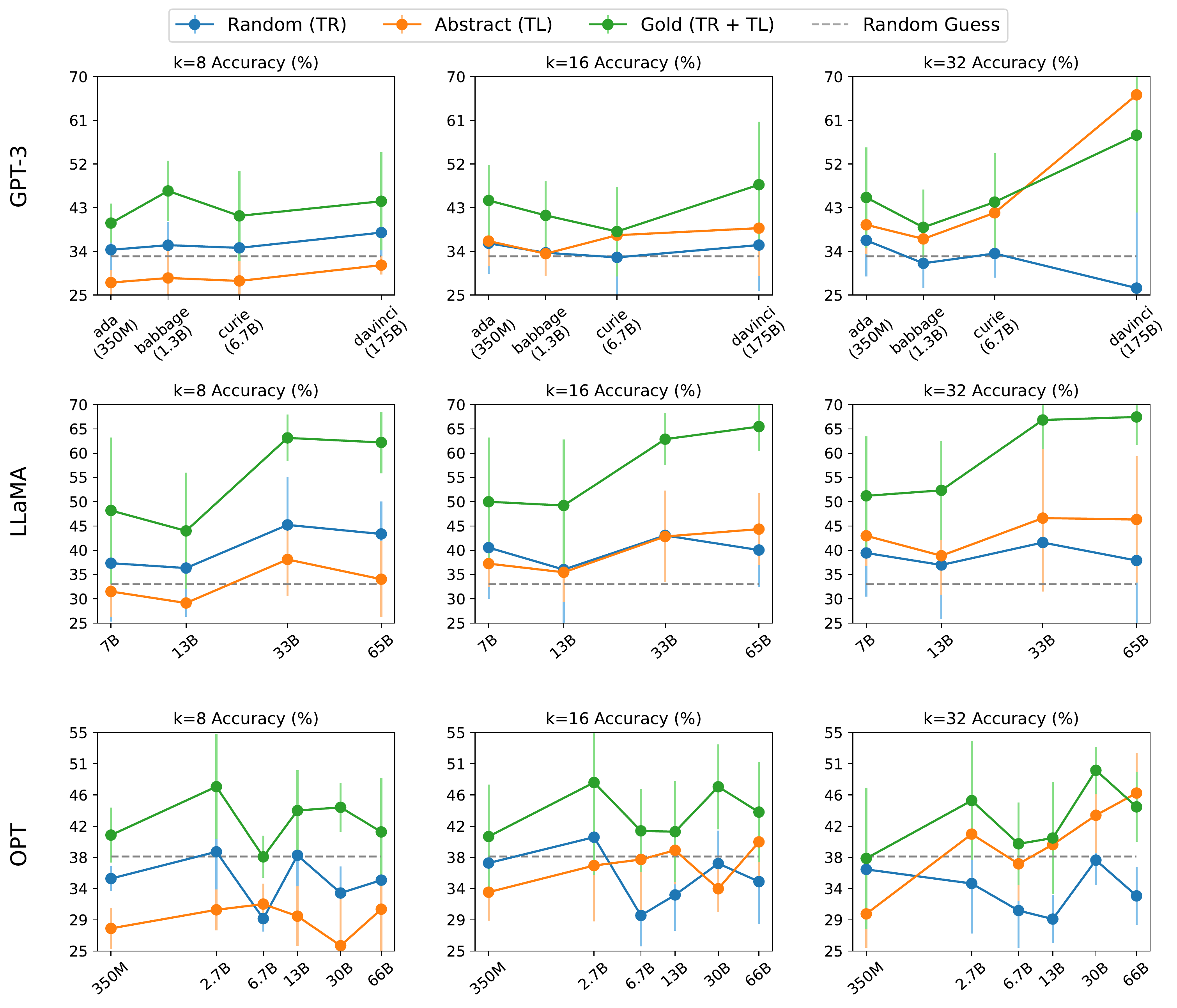}
    \caption{Average performance of \textbf{topic/stance classification} datasets.%
    } 
    \label{fig:topic_detection}
\end{figure*}

Figure~\ref{fig:sentiment_analysis}, Figure~\ref{fig:NLI}, Figure~\ref{fig:toxicity_detection}, and Figure~\ref{fig:topic_detection}
show task-type breakdown results. 
Though individual task-type results are more noisy, we 
can make a similar observation compared to the  main result --
\abstractl{} (TL) scales better with sizes and numbers of examples compared to \randoml{} (TR). 

\newpage

\newcommand{\bsl}{\makebox[0pt][r]{\raisebox{0.05em}{$\bigstar\,$}}}

\setlength{\tabcolsep}{0.2cm}
\begin{table*}[ht]
    \centering
    \resizebox{0.98\textwidth}{!}{
        \begin{tabular}{lcccccccccccccccc}
\toprule
&\multicolumn{3}{c}{\tf{tweet\_eval\_hate}} && \multicolumn{3}{c}{\tf{tweet\_eval\_atheism}} && \multicolumn{3}{c}{\tf{tweet\_eval\_feminist}} && \multicolumn{3}{c}{\tf{sick}} \\
&Random & Abstract & Gold && Random & Abstract & Gold && Random & Abstract & Gold && Random & Abstract & Gold \\
\midrule\
\texttt{ada} & 0.52 & 0.51 & 0.54&& 0.45 & 0.23 & 0.4&& 0.4 & 0.38 & 0.41&& 0.44 & 0.34 & 0.43 \\
\texttt{babbage} & 0.51 & 0.52 & 0.54&& 0.38 & 0.37 & 0.43&& 0.46 & 0.29 & 0.49&& 0.53 & 0.34 & 0.57 \\
\texttt{curie} & 0.55 & 0.54 & 0.6&& 0.28 & 0.33 & 0.32&& 0.39 & 0.32 & 0.4&& 0.56 & 0.36 & 0.56 \\
\texttt{davinci} & 0.56 & 0.55 & 0.59&& 0.34 & 0.33 & 0.33&& 0.4 & 0.4 & 0.38&& 0.4 & 0.39 & 0.44 \\
\toprule
&\multicolumn{3}{c}{\tf{financial\_phrasebank}} && \multicolumn{3}{c}{\tf{ethos\_race}} && \multicolumn{3}{c}{\tf{ethos\_gender}} && \multicolumn{3}{c}{\tf{ethos\_religion}} \\
&Random & Abstract & Gold && Random & Abstract & Gold && Random & Abstract & Gold && Random & Abstract & Gold \\
\midrule\
\texttt{ada} & 0.23 & 0.4 & 0.4&& 0.38 & 0.41 & 0.44&& 0.34 & 0.43 & 0.56&& 0.39 & 0.64 & 0.62 \\
\texttt{babbage} & 0.37 & 0.43 & 0.46&& 0.29 & 0.49 & 0.53&& 0.34 & 0.57 & 0.45&& 0.39 & 0.55 & 0.54 \\
\texttt{curie} & 0.33 & 0.32 & 0.39&& 0.32 & 0.4 & 0.56&& 0.36 & 0.56 & 0.54&& 0.42 & 0.63 & 0.53 \\
\texttt{davinci} & 0.33 & 0.33 & 0.4&& 0.4 & 0.38 & 0.4&& 0.39 & 0.44 & 0.4&& 0.56 & 0.44 & 0.52 \\
\toprule
&\multicolumn{3}{c}{\tf{ethos\_national\_origin}} && \multicolumn{3}{c}{\tf{snli}} && \multicolumn{3}{c}{\tf{sst2}} && \multicolumn{3}{c}{\tf{trec}} \\
&Random & Abstract & Gold && Random & Abstract & Gold && Random & Abstract & Gold && Random & Abstract & Gold \\
\midrule\
\texttt{ada} & 0.41 & 0.44 & 0.34&& 0.43 & 0.56 & 0.39&& 0.64 & 0.62 & 0.52&& 0.71 & 0.62 & 0.57 \\
\texttt{babbage} & 0.49 & 0.53 & 0.34&& 0.57 & 0.45 & 0.39&& 0.55 & 0.54 & 0.51&& 0.52 & 0.58 & 0.56 \\
\texttt{curie} & 0.4 & 0.56 & 0.36&& 0.56 & 0.54 & 0.42&& 0.63 & 0.53 & 0.52&& 0.6 & 0.48 & 0.54 \\
\texttt{davinci} & 0.38 & 0.4 & 0.39&& 0.44 & 0.4 & 0.56&& 0.44 & 0.52 & 0.51&& 0.54 & 0.47 & 0.52 \\
\toprule
&\multicolumn{3}{c}{\tf{rte}} && \multicolumn{3}{c}{\tf{wnli}} && \multicolumn{3}{c}{\tf{mrpc}} && \multicolumn{3}{c}{\tf{poem}} \\
&Random & Abstract & Gold && Random & Abstract & Gold && Random & Abstract & Gold && Random & Abstract & Gold \\
\midrule\
\texttt{ada} & 0.56 & 0.39 & 0.64&& 0.62 & 0.52 & 0.71&& 0.62 & 0.57 & 0.76&& 0.68 & 0.54 & 0.74 \\
\texttt{babbage} & 0.45 & 0.39 & 0.55&& 0.54 & 0.51 & 0.52&& 0.58 & 0.56 & 0.62&& 0.63 & 0.51 & 0.61 \\
\texttt{curie} & 0.54 & 0.42 & 0.63&& 0.53 & 0.52 & 0.6&& 0.48 & 0.54 & 0.55&& 0.56 & 0.6 & 0.54 \\
\texttt{davinci} & 0.4 & 0.56 & 0.44&& 0.52 & 0.51 & 0.54&& 0.47 & 0.52 & 0.5&& 0.48 & 0.53 & 0.62 \\

\bottomrule

        \end{tabular}
    }
    \caption{
    Single dataset accuracies across the GPT-3 model family, using 8 examples.
    }
    \label{tab:gpt_8}
    \vspace{-3pt}
\end{table*}

\setlength{\tabcolsep}{0.2cm}
\begin{table*}[ht]
    \centering
    \resizebox{0.98\textwidth}{!}{
        \begin{tabular}{lcccccccccccccccc}
\toprule
&\multicolumn{3}{c}{\tf{tweet\_eval\_hate}} && \multicolumn{3}{c}{\tf{tweet\_eval\_atheism}} && \multicolumn{3}{c}{\tf{tweet\_eval\_feminist}} && \multicolumn{3}{c}{\tf{sick}} \\
&Random & Abstract & Gold && Random & Abstract & Gold && Random & Abstract & Gold && Random & Abstract & Gold \\
\midrule\
\texttt{ada} & 0.51 & 0.51 & 0.52&& 0.44 & 0.37 & 0.48&& 0.4 & 0.42 & 0.41&& 0.37 & 0.44 & 0.44 \\
\texttt{babbage} & 0.48 & 0.54 & 0.55&& 0.36 & 0.41 & 0.31&& 0.44 & 0.33 & 0.48&& 0.54 & 0.38 & 0.54 \\
\texttt{curie} & 0.54 & 0.58 & 0.62&& 0.28 & 0.48 & 0.3&& 0.33 & 0.38 & 0.32&& 0.56 & 0.41 & 0.56 \\
\texttt{davinci} & 0.56 & 0.6 & 0.64&& 0.34 & 0.42 & 0.39&& 0.29 & 0.44 & 0.38&& 0.46 & 0.49 & 0.49 \\
\toprule
&\multicolumn{3}{c}{\tf{financial\_phrasebank}} && \multicolumn{3}{c}{\tf{ethos\_race}} && \multicolumn{3}{c}{\tf{ethos\_gender}} && \multicolumn{3}{c}{\tf{ethos\_religion}} \\
&Random & Abstract & Gold && Random & Abstract & Gold && Random & Abstract & Gold && Random & Abstract & Gold \\
\midrule\
\texttt{ada} & 0.37 & 0.48 & 0.4&& 0.42 & 0.41 & 0.37&& 0.44 & 0.44 & 0.54&& 0.53 & 0.67 & 0.68 \\
\texttt{babbage} & 0.41 & 0.31 & 0.44&& 0.33 & 0.48 & 0.54&& 0.38 & 0.54 & 0.43&& 0.53 & 0.63 & 0.56 \\
\texttt{curie} & 0.48 & 0.3 & 0.33&& 0.38 & 0.32 & 0.56&& 0.41 & 0.56 & 0.5&& 0.55 & 0.71 & 0.54 \\
\texttt{davinci} & 0.42 & 0.39 & 0.29&& 0.44 & 0.38 & 0.46&& 0.49 & 0.49 & 0.38&& 0.63 & 0.49 & 0.51 \\
\toprule
&\multicolumn{3}{c}{\tf{ethos\_national\_origin}} && \multicolumn{3}{c}{\tf{snli}} && \multicolumn{3}{c}{\tf{sst2}} && \multicolumn{3}{c}{\tf{trec}} \\
&Random & Abstract & Gold && Random & Abstract & Gold && Random & Abstract & Gold && Random & Abstract & Gold \\
\midrule\
\texttt{ada} & 0.41 & 0.37 & 0.44&& 0.44 & 0.54 & 0.53&& 0.67 & 0.68 & 0.52&& 0.76 & 0.68 & 0.52 \\
\texttt{babbage} & 0.48 & 0.54 & 0.38&& 0.54 & 0.43 & 0.53&& 0.63 & 0.56 & 0.53&& 0.61 & 0.58 & 0.54 \\
\texttt{curie} & 0.32 & 0.56 & 0.41&& 0.56 & 0.5 & 0.55&& 0.71 & 0.54 & 0.56&& 0.55 & 0.49 & 0.55 \\
\texttt{davinci} & 0.38 & 0.46 & 0.49&& 0.49 & 0.38 & 0.63&& 0.49 & 0.51 & 0.6&& 0.56 & 0.5 & 0.57 \\
\toprule
&\multicolumn{3}{c}{\tf{rte}} && \multicolumn{3}{c}{\tf{wnli}} && \multicolumn{3}{c}{\tf{mrpc}} && \multicolumn{3}{c}{\tf{poem}} \\
&Random & Abstract & Gold && Random & Abstract & Gold && Random & Abstract & Gold && Random & Abstract & Gold \\
\midrule\
\texttt{ada} & 0.54 & 0.53 & 0.67&& 0.68 & 0.52 & 0.76&& 0.68 & 0.52 & 0.75&& 0.69 & 0.57 & 0.77 \\
\texttt{babbage} & 0.43 & 0.53 & 0.63&& 0.56 & 0.53 & 0.61&& 0.58 & 0.54 & 0.61&& 0.55 & 0.54 & 0.62 \\
\texttt{curie} & 0.5 & 0.55 & 0.71&& 0.54 & 0.56 & 0.55&& 0.49 & 0.55 & 0.59&& 0.49 & 0.55 & 0.59 \\
\texttt{davinci} & 0.38 & 0.63 & 0.49&& 0.51 & 0.6 & 0.56&& 0.5 & 0.57 & 0.59&& 0.54 & 0.67 & 0.63 \\

\bottomrule

        \end{tabular}
    }
    \caption{
    Single dataset accuracies across the GPT-3 model family, using 16 examples.
    }
    \label{tab:gpt_16}
    \vspace{-3pt}
\end{table*}

\setlength{\tabcolsep}{0.2cm}
\begin{table*}[ht]
    \centering
    \resizebox{0.98\textwidth}{!}{
        \begin{tabular}{lcccccccccccccccc}
\toprule
&\multicolumn{3}{c}{\tf{tweet\_eval\_hate}} && \multicolumn{3}{c}{\tf{tweet\_eval\_atheism}} && \multicolumn{3}{c}{\tf{tweet\_eval\_feminist}} && \multicolumn{3}{c}{\tf{sick}} \\
&Random & Abstract & Gold && Random & Abstract & Gold && Random & Abstract & Gold && Random & Abstract & Gold \\
\midrule\
\texttt{ada} & 0.48 & 0.52 & 0.53&& 0.4 & 0.37 & 0.42&& 0.41 & 0.38 & 0.42&& 0.24 & 0.45 & 0.27 \\
\texttt{babbage} & 0.53 & 0.58 & 0.52&& 0.32 & 0.38 & 0.35&& 0.42 & 0.35 & 0.38&& 0.44 & 0.4 & 0.5 \\
\texttt{curie} & 0.54 & 0.59 & 0.66&& 0.26 & 0.47 & 0.31&& 0.38 & 0.4 & 0.43&& 0.57 & 0.41 & 0.57 \\
\texttt{davinci} & 0.57 & 0.64 & 0.66&& 0.29 & 0.51 & 0.37&& 0.28 & 0.49 & 0.37&& 0.43 & 0.52 & 0.49 \\
\toprule
&\multicolumn{3}{c}{\tf{financial\_phrasebank}} && \multicolumn{3}{c}{\tf{ethos\_race}} && \multicolumn{3}{c}{\tf{ethos\_gender}} && \multicolumn{3}{c}{\tf{ethos\_religion}} \\
&Random & Abstract & Gold && Random & Abstract & Gold && Random & Abstract & Gold && Random & Abstract & Gold \\
\midrule\
\texttt{ada} & 0.37 & 0.42 & 0.41&& 0.38 & 0.42 & 0.24&& 0.45 & 0.27 & 0.55&& 0.56 & 0.69 & 0.66 \\
\texttt{babbage} & 0.38 & 0.35 & 0.42&& 0.35 & 0.38 & 0.44&& 0.4 & 0.5 & 0.51&& 0.58 & 0.65 & 0.51 \\
\texttt{curie} & 0.47 & 0.31 & 0.38&& 0.4 & 0.43 & 0.57&& 0.41 & 0.57 & 0.52&& 0.56 & 0.71 & 0.51 \\
\texttt{davinci} & 0.51 & 0.37 & 0.28&& 0.49 & 0.37 & 0.43&& 0.52 & 0.49 & 0.35&& 0.68 & 0.5 & 0.63 \\
\toprule
&\multicolumn{3}{c}{\tf{ethos\_national\_origin}} && \multicolumn{3}{c}{\tf{snli}} && \multicolumn{3}{c}{\tf{sst2}} && \multicolumn{3}{c}{\tf{trec}} \\
&Random & Abstract & Gold && Random & Abstract & Gold && Random & Abstract & Gold && Random & Abstract & Gold \\
\midrule\
\texttt{ada} & 0.42 & 0.24 & 0.45&& 0.27 & 0.55 & 0.56&& 0.69 & 0.66 & 0.55&& 0.73 & 0.69 & 0.61 \\
\texttt{babbage} & 0.38 & 0.44 & 0.4&& 0.5 & 0.51 & 0.58&& 0.65 & 0.51 & 0.57&& 0.63 & 0.6 & 0.59 \\
\texttt{curie} & 0.43 & 0.57 & 0.41&& 0.57 & 0.52 & 0.56&& 0.71 & 0.51 & 0.59&& 0.65 & 0.5 & 0.61 \\
\texttt{davinci} & 0.37 & 0.43 & 0.52&& 0.49 & 0.35 & 0.68&& 0.5 & 0.63 & 0.6&& 0.63 & 0.51 & 0.62 \\
\toprule
&\multicolumn{3}{c}{\tf{rte}} && \multicolumn{3}{c}{\tf{wnli}} && \multicolumn{3}{c}{\tf{mrpc}} && \multicolumn{3}{c}{\tf{poem}} \\
&Random & Abstract & Gold && Random & Abstract & Gold && Random & Abstract & Gold && Random & Abstract & Gold \\
\midrule\
\texttt{ada} & 0.55 & 0.56 & 0.69&& 0.66 & 0.55 & 0.73&& 0.69 & 0.61 & 0.73&& 0.65 & 0.63 & 0.77 \\
\texttt{babbage} & 0.51 & 0.58 & 0.65&& 0.51 & 0.57 & 0.63&& 0.6 & 0.59 & 0.64&& 0.57 & 0.56 & 0.65 \\
\texttt{curie} & 0.52 & 0.56 & 0.71&& 0.51 & 0.59 & 0.65&& 0.5 & 0.61 & 0.63&& 0.44 & 0.61 & 0.69 \\
\texttt{davinci} & 0.35 & 0.68 & 0.5&& 0.63 & 0.6 & 0.63&& 0.51 & 0.62 & 0.65&& 0.6 & 0.7 & 0.71 \\

\bottomrule

        \end{tabular}
    }
    \caption{
    Single dataset accuracies across the GPT-3 model family, using 32 examples.
    }
    \label{tab:gpt_32}
    \vspace{-3pt}
\end{table*}

\setlength{\tabcolsep}{0.2cm}
\begin{table*}[ht]
    \centering
    \resizebox{0.98\textwidth}{!}{
        \begin{tabular}{lcccccccccccccccc}
\toprule
&\multicolumn{3}{c}{\tf{tweet\_eval\_hate}} && \multicolumn{3}{c}{\tf{tweet\_eval\_atheism}} && \multicolumn{3}{c}{\tf{tweet\_eval\_feminist}} && \multicolumn{3}{c}{\tf{sick}} \\
&Random & Abstract & Gold && Random & Abstract & Gold && Random & Abstract & Gold && Random & Abstract & Gold \\
\midrule\
\texttt{OPT-350M} & 0.49 & 0.51 & 0.53&& 0.43 & 0.34 & 0.48&& 0.41 & 0.31 & 0.45&& 0.33 & 0.34 & 0.29 \\
\texttt{OPT-2.7B} & 0.52 & 0.55 & 0.56&& 0.43 & 0.36 & 0.45&& 0.47 & 0.34 & 0.5&& 0.52 & 0.34 & 0.55 \\
\texttt{OPT-6.7B} & 0.53 & 0.53 & 0.57&& 0.26 & 0.33 & 0.27&& 0.33 & 0.39 & 0.36&& 0.46 & 0.36 & 0.48 \\
\texttt{OPT-13B} & 0.55 & 0.52 & 0.61&& 0.4 & 0.35 & 0.4&& 0.49 & 0.35 & 0.47&& 0.36 & 0.3 & 0.37 \\
\texttt{OPT-30B} & 0.52 & 0.54 & 0.55&& 0.28 & 0.24 & 0.35&& 0.4 & 0.34 & 0.46&& 0.53 & 0.31 & 0.55 \\
\texttt{OPT-66B} & 0.52 & 0.55 & 0.53&& 0.29 & 0.38 & 0.32&& 0.44 & 0.37 & 0.42&& 0.44 & 0.36 & 0.47 \\
\toprule
&\multicolumn{3}{c}{\tf{financial\_phrasebank}} && \multicolumn{3}{c}{\tf{ethos\_race}} && \multicolumn{3}{c}{\tf{ethos\_gender}} && \multicolumn{3}{c}{\tf{ethos\_religion}} \\
&Random & Abstract & Gold && Random & Abstract & Gold && Random & Abstract & Gold && Random & Abstract & Gold \\
\midrule\
\texttt{OPT-350M} & 0.34 & 0.48 & 0.41&& 0.31 & 0.45 & 0.33&& 0.34 & 0.29 & 0.48&& 0.36 & 0.48 & 0.6 \\
\texttt{OPT-2.7B} & 0.36 & 0.45 & 0.47&& 0.34 & 0.5 & 0.52&& 0.34 & 0.55 & 0.54&& 0.42 & 0.56 & 0.49 \\
\texttt{OPT-6.7B} & 0.33 & 0.27 & 0.33&& 0.39 & 0.36 & 0.46&& 0.36 & 0.48 & 0.63&& 0.44 & 0.74 & 0.55 \\
\texttt{OPT-13B} & 0.35 & 0.4 & 0.49&& 0.35 & 0.47 & 0.36&& 0.3 & 0.37 & 0.59&& 0.44 & 0.69 & 0.62 \\
\texttt{OPT-30B} & 0.24 & 0.35 & 0.4&& 0.34 & 0.46 & 0.53&& 0.31 & 0.55 & 0.56&& 0.43 & 0.61 & 0.44 \\
\texttt{OPT-66B} & 0.38 & 0.32 & 0.44&& 0.37 & 0.42 & 0.44&& 0.36 & 0.47 & 0.33&& 0.44 & 0.46 & 0.45 \\
\toprule
&\multicolumn{3}{c}{\tf{ethos\_national\_origin}} && \multicolumn{3}{c}{\tf{snli}} && \multicolumn{3}{c}{\tf{sst2}} && \multicolumn{3}{c}{\tf{trec}} \\
&Random & Abstract & Gold && Random & Abstract & Gold && Random & Abstract & Gold && Random & Abstract & Gold \\
\midrule\
\texttt{OPT-350M} & 0.45 & 0.33 & 0.34&& 0.29 & 0.48 & 0.36&& 0.48 & 0.6 & 0.49&& 0.66 & 0.65 & 0.51 \\
\texttt{OPT-2.7B} & 0.5 & 0.52 & 0.34&& 0.55 & 0.54 & 0.42&& 0.56 & 0.49 & 0.53&& 0.49 & 0.51 & 0.56 \\
\texttt{OPT-6.7B} & 0.36 & 0.46 & 0.36&& 0.48 & 0.63 & 0.44&& 0.74 & 0.55 & 0.54&& 0.59 & 0.53 & 0.57 \\
\texttt{OPT-13B} & 0.47 & 0.36 & 0.3&& 0.37 & 0.59 & 0.44&& 0.69 & 0.62 & 0.53&& 0.63 & 0.55 & 0.53 \\
\texttt{OPT-30B} & 0.46 & 0.53 & 0.31&& 0.55 & 0.56 & 0.43&& 0.61 & 0.44 & 0.49&& 0.42 & 0.46 & 0.57 \\
\texttt{OPT-66B} & 0.42 & 0.44 & 0.36&& 0.47 & 0.33 & 0.44&& 0.46 & 0.45 & 0.55&& 0.53 & 0.44 & 0.55 \\
\toprule
&\multicolumn{3}{c}{\tf{rte}} && \multicolumn{3}{c}{\tf{wnli}} && \multicolumn{3}{c}{\tf{mrpc}} && \multicolumn{3}{c}{\tf{poem}} \\
&Random & Abstract & Gold && Random & Abstract & Gold && Random & Abstract & Gold && Random & Abstract & Gold \\
\midrule\
\texttt{OPT-350M} & 0.48 & 0.36 & 0.48&& 0.6 & 0.49 & 0.66&& 0.65 & 0.51 & 0.71&& 0.66 & 0.53 & 0.73 \\
\texttt{OPT-2.7B} & 0.54 & 0.42 & 0.56&& 0.49 & 0.53 & 0.49&& 0.51 & 0.56 & 0.52&& 0.48 & 0.51 & 0.5 \\
\texttt{OPT-6.7B} & 0.63 & 0.44 & 0.74&& 0.55 & 0.54 & 0.59&& 0.53 & 0.57 & 0.61&& 0.53 & 0.52 & 0.62 \\
\texttt{OPT-13B} & 0.59 & 0.44 & 0.69&& 0.62 & 0.53 & 0.63&& 0.55 & 0.53 & 0.61&& 0.55 & 0.52 & 0.62 \\
\texttt{OPT-30B} & 0.56 & 0.43 & 0.61&& 0.44 & 0.49 & 0.42&& 0.46 & 0.57 & 0.47&& 0.46 & 0.51 & 0.46 \\
\texttt{OPT-66B} & 0.33 & 0.44 & 0.46&& 0.45 & 0.55 & 0.53&& 0.44 & 0.55 & 0.38&& 0.49 & 0.55 & 0.56 \\

\bottomrule

        \end{tabular}
    }
    \caption{
    Single dataset accuracies across the OPT model family, using 8 examples.
    }
    \label{tab:opt_8}
    \vspace{-3pt}
\end{table*}

\setlength{\tabcolsep}{0.2cm}
\begin{table*}[ht]
    \centering
    \resizebox{0.98\textwidth}{!}{
        \begin{tabular}{lcccccccccccccccc}
\toprule
&\multicolumn{3}{c}{\tf{tweet\_eval\_hate}} && \multicolumn{3}{c}{\tf{tweet\_eval\_atheism}} && \multicolumn{3}{c}{\tf{tweet\_eval\_feminist}} && \multicolumn{3}{c}{\tf{sick}} \\
&Random & Abstract & Gold && Random & Abstract & Gold && Random & Abstract & Gold && Random & Abstract & Gold \\
\midrule\
\texttt{OPT-350M} & 0.52 & 0.53 & 0.55&& 0.47 & 0.37 & 0.49&& 0.42 & 0.42 & 0.44&& 0.33 & 0.36 & 0.35 \\
\texttt{OPT-2.7B} & 0.52 & 0.56 & 0.58&& 0.44 & 0.44 & 0.47&& 0.51 & 0.39 & 0.46&& 0.55 & 0.39 & 0.57 \\
\texttt{OPT-6.7B} & 0.52 & 0.57 & 0.57&& 0.22 & 0.39 & 0.28&& 0.39 & 0.43 & 0.41&& 0.48 & 0.42 & 0.54 \\
\texttt{OPT-13B} & 0.58 & 0.54 & 0.62&& 0.32 & 0.44 & 0.38&& 0.41 & 0.39 & 0.41&& 0.36 & 0.4 & 0.36 \\
\texttt{OPT-30B} & 0.51 & 0.57 & 0.57&& 0.34 & 0.4 & 0.35&& 0.41 & 0.32 & 0.5&& 0.55 & 0.45 & 0.56 \\
\texttt{OPT-66B} & 0.5 & 0.57 & 0.54&& 0.25 & 0.47 & 0.31&& 0.47 & 0.44 & 0.48&& 0.49 & 0.38 & 0.51 \\
\toprule
&\multicolumn{3}{c}{\tf{financial\_phrasebank}} && \multicolumn{3}{c}{\tf{ethos\_race}} && \multicolumn{3}{c}{\tf{ethos\_gender}} && \multicolumn{3}{c}{\tf{ethos\_religion}} \\
&Random & Abstract & Gold && Random & Abstract & Gold && Random & Abstract & Gold && Random & Abstract & Gold \\
\midrule\
\texttt{OPT-350M} & 0.37 & 0.49 & 0.42&& 0.42 & 0.44 & 0.33&& 0.36 & 0.35 & 0.45&& 0.4 & 0.47 & 0.65 \\
\texttt{OPT-2.7B} & 0.44 & 0.47 & 0.51&& 0.39 & 0.46 & 0.55&& 0.39 & 0.57 & 0.53&& 0.5 & 0.58 & 0.45 \\
\texttt{OPT-6.7B} & 0.39 & 0.28 & 0.39&& 0.43 & 0.41 & 0.48&& 0.42 & 0.54 & 0.66&& 0.53 & 0.8 & 0.59 \\
\texttt{OPT-13B} & 0.44 & 0.38 & 0.41&& 0.39 & 0.41 & 0.36&& 0.4 & 0.36 & 0.6&& 0.53 & 0.72 & 0.54 \\
\texttt{OPT-30B} & 0.4 & 0.35 & 0.41&& 0.32 & 0.5 & 0.55&& 0.45 & 0.56 & 0.56&& 0.52 & 0.64 & 0.35 \\
\texttt{OPT-66B} & 0.47 & 0.31 & 0.47&& 0.44 & 0.48 & 0.49&& 0.38 & 0.51 & 0.3&& 0.57 & 0.49 & 0.44 \\
\toprule
&\multicolumn{3}{c}{\tf{ethos\_national\_origin}} && \multicolumn{3}{c}{\tf{snli}} && \multicolumn{3}{c}{\tf{sst2}} && \multicolumn{3}{c}{\tf{trec}} \\
&Random & Abstract & Gold && Random & Abstract & Gold && Random & Abstract & Gold && Random & Abstract & Gold \\
\midrule\
\texttt{OPT-350M} & 0.44 & 0.33 & 0.36&& 0.35 & 0.45 & 0.4&& 0.47 & 0.65 & 0.52&& 0.71 & 0.71 & 0.52 \\
\texttt{OPT-2.7B} & 0.46 & 0.55 & 0.39&& 0.57 & 0.53 & 0.5&& 0.58 & 0.45 & 0.59&& 0.47 & 0.41 & 0.62 \\
\texttt{OPT-6.7B} & 0.41 & 0.48 & 0.42&& 0.54 & 0.66 & 0.53&& 0.8 & 0.59 & 0.56&& 0.71 & 0.62 & 0.61 \\
\texttt{OPT-13B} & 0.41 & 0.36 & 0.4&& 0.36 & 0.6 & 0.53&& 0.72 & 0.54 & 0.53&& 0.62 & 0.5 & 0.55 \\
\texttt{OPT-30B} & 0.5 & 0.55 & 0.45&& 0.56 & 0.56 & 0.52&& 0.64 & 0.35 & 0.57&& 0.43 & 0.38 & 0.63 \\
\texttt{OPT-66B} & 0.48 & 0.49 & 0.38&& 0.51 & 0.3 & 0.57&& 0.49 & 0.44 & 0.59&& 0.51 & 0.4 & 0.6 \\
\toprule
&\multicolumn{3}{c}{\tf{rte}} && \multicolumn{3}{c}{\tf{wnli}} && \multicolumn{3}{c}{\tf{mrpc}} && \multicolumn{3}{c}{\tf{poem}} \\
&Random & Abstract & Gold && Random & Abstract & Gold && Random & Abstract & Gold && Random & Abstract & Gold \\
\midrule\
\texttt{OPT-350M} & 0.45 & 0.4 & 0.47&& 0.65 & 0.52 & 0.71&& 0.71 & 0.52 & 0.76&& 0.73 & 0.51 & 0.76 \\
\texttt{OPT-2.7B} & 0.53 & 0.5 & 0.58&& 0.45 & 0.59 & 0.47&& 0.41 & 0.62 & 0.52&& 0.45 & 0.54 & 0.54 \\
\texttt{OPT-6.7B} & 0.66 & 0.53 & 0.8&& 0.59 & 0.56 & 0.71&& 0.62 & 0.61 & 0.69&& 0.64 & 0.61 & 0.74 \\
\texttt{OPT-13B} & 0.6 & 0.53 & 0.72&& 0.54 & 0.53 & 0.62&& 0.5 & 0.55 & 0.58&& 0.55 & 0.53 & 0.58 \\
\texttt{OPT-30B} & 0.56 & 0.52 & 0.64&& 0.35 & 0.57 & 0.43&& 0.38 & 0.63 & 0.5&& 0.41 & 0.59 & 0.51 \\
\texttt{OPT-66B} & 0.3 & 0.57 & 0.49&& 0.44 & 0.59 & 0.51&& 0.4 & 0.6 & 0.46&& 0.46 & 0.59 & 0.55 \\

\bottomrule

        \end{tabular}
    }
    \caption{
    Single dataset accuracies across the OPT model family, using 16 examples.
    }
    \label{tab:opt_16}
    \vspace{-3pt}
\end{table*}

\setlength{\tabcolsep}{0.2cm}
\begin{table*}[ht]
    \centering
    \resizebox{0.98\textwidth}{!}{
        \begin{tabular}{lcccccccccccccccc}
\toprule
&\multicolumn{3}{c}{\tf{tweet\_eval\_hate}} && \multicolumn{3}{c}{\tf{tweet\_eval\_atheism}} && \multicolumn{3}{c}{\tf{tweet\_eval\_feminist}} && \multicolumn{3}{c}{\tf{sick}} \\
&Random & Abstract & Gold && Random & Abstract & Gold && Random & Abstract & Gold && Random & Abstract & Gold \\
\midrule\
\texttt{OPT-350M} & 0.53 & 0.53 & 0.55&& 0.42 & 0.35 & 0.42&& 0.43 & 0.33 & 0.4&& 0.36 & 0.34 & 0.35 \\
\texttt{OPT-2.7B} & 0.51 & 0.59 & 0.59&& 0.31 & 0.42 & 0.42&& 0.43 & 0.39 & 0.42&& 0.53 & 0.4 & 0.57 \\
\texttt{OPT-6.7B} & 0.55 & 0.59 & 0.6&& 0.26 & 0.29 & 0.24&& 0.4 & 0.39 & 0.42&& 0.49 & 0.44 & 0.53 \\
\texttt{OPT-13B} & 0.56 & 0.58 & 0.59&& 0.25 & 0.45 & 0.36&& 0.39 & 0.38 & 0.42&& 0.4 & 0.38 & 0.37 \\
\texttt{OPT-30B} & 0.52 & 0.59 & 0.57&& 0.32 & 0.47 & 0.42&& 0.47 & 0.42 & 0.47&& 0.54 & 0.45 & 0.6 \\
\texttt{OPT-66B} & 0.48 & 0.58 & 0.51&& 0.27 & 0.5 & 0.26&& 0.4 & 0.46 & 0.5&& 0.45 & 0.43 & 0.47 \\
\toprule
&\multicolumn{3}{c}{\tf{financial\_phrasebank}} && \multicolumn{3}{c}{\tf{ethos\_race}} && \multicolumn{3}{c}{\tf{ethos\_gender}} && \multicolumn{3}{c}{\tf{ethos\_religion}} \\
&Random & Abstract & Gold && Random & Abstract & Gold && Random & Abstract & Gold && Random & Abstract & Gold \\
\midrule\
\texttt{OPT-350M} & 0.35 & 0.42 & 0.43&& 0.33 & 0.4 & 0.36&& 0.34 & 0.35 & 0.44&& 0.38 & 0.44 & 0.67 \\
\texttt{OPT-2.7B} & 0.42 & 0.42 & 0.43&& 0.39 & 0.42 & 0.53&& 0.4 & 0.57 & 0.51&& 0.56 & 0.58 & 0.46 \\
\texttt{OPT-6.7B} & 0.29 & 0.24 & 0.4&& 0.39 & 0.42 & 0.49&& 0.44 & 0.53 & 0.68&& 0.61 & 0.82 & 0.63 \\
\texttt{OPT-13B} & 0.45 & 0.36 & 0.39&& 0.38 & 0.42 & 0.4&& 0.38 & 0.37 & 0.61&& 0.6 & 0.72 & 0.48 \\
\texttt{OPT-30B} & 0.47 & 0.42 & 0.47&& 0.42 & 0.47 & 0.54&& 0.45 & 0.6 & 0.57&& 0.57 & 0.7 & 0.4 \\
\texttt{OPT-66B} & 0.5 & 0.26 & 0.4&& 0.46 & 0.5 & 0.45&& 0.43 & 0.47 & 0.37&& 0.64 & 0.57 & 0.41 \\
\toprule
&\multicolumn{3}{c}{\tf{ethos\_national\_origin}} && \multicolumn{3}{c}{\tf{snli}} && \multicolumn{3}{c}{\tf{sst2}} && \multicolumn{3}{c}{\tf{trec}} \\
&Random & Abstract & Gold && Random & Abstract & Gold && Random & Abstract & Gold && Random & Abstract & Gold \\
\midrule\
\texttt{OPT-350M} & 0.4 & 0.36 & 0.34&& 0.35 & 0.44 & 0.38&& 0.44 & 0.67 & 0.51&& 0.73 & 0.71 & 0.51 \\
\texttt{OPT-2.7B} & 0.42 & 0.53 & 0.4&& 0.57 & 0.51 & 0.56&& 0.58 & 0.46 & 0.55&& 0.49 & 0.43 & 0.6 \\
\texttt{OPT-6.7B} & 0.42 & 0.49 & 0.44&& 0.53 & 0.68 & 0.61&& 0.82 & 0.63 & 0.65&& 0.74 & 0.62 & 0.65 \\
\texttt{OPT-13B} & 0.42 & 0.4 & 0.38&& 0.37 & 0.61 & 0.6&& 0.72 & 0.48 & 0.56&& 0.57 & 0.44 & 0.64 \\
\texttt{OPT-30B} & 0.47 & 0.54 & 0.45&& 0.6 & 0.57 & 0.57&& 0.7 & 0.4 & 0.55&& 0.42 & 0.36 & 0.66 \\
\texttt{OPT-66B} & 0.5 & 0.45 & 0.43&& 0.47 & 0.37 & 0.64&& 0.57 & 0.41 & 0.63&& 0.52 & 0.36 & 0.67 \\
\toprule
&\multicolumn{3}{c}{\tf{rte}} && \multicolumn{3}{c}{\tf{wnli}} && \multicolumn{3}{c}{\tf{mrpc}} && \multicolumn{3}{c}{\tf{poem}} \\
&Random & Abstract & Gold && Random & Abstract & Gold && Random & Abstract & Gold && Random & Abstract & Gold \\
\midrule\
\texttt{OPT-350M} & 0.44 & 0.38 & 0.44&& 0.67 & 0.51 & 0.73&& 0.71 & 0.51 & 0.77&& 0.74 & 0.52 & 0.79 \\
\texttt{OPT-2.7B} & 0.51 & 0.56 & 0.58&& 0.46 & 0.55 & 0.49&& 0.43 & 0.6 & 0.54&& 0.41 & 0.56 & 0.48 \\
\texttt{OPT-6.7B} & 0.68 & 0.61 & 0.82&& 0.63 & 0.65 & 0.74&& 0.62 & 0.65 & 0.78&& 0.65 & 0.64 & 0.77 \\
\texttt{OPT-13B} & 0.61 & 0.6 & 0.72&& 0.48 & 0.56 & 0.57&& 0.44 & 0.64 & 0.5&& 0.45 & 0.53 & 0.5 \\
\texttt{OPT-30B} & 0.57 & 0.57 & 0.7&& 0.4 & 0.55 & 0.42&& 0.36 & 0.66 & 0.46&& 0.4 & 0.71 & 0.54 \\
\texttt{OPT-66B} & 0.37 & 0.64 & 0.57&& 0.41 & 0.63 & 0.52&& 0.36 & 0.67 & 0.49&& 0.4 & 0.69 & 0.56 \\

\bottomrule

        \end{tabular}
    }
    \caption{
    Single dataset accuracies across the OPT model family, using 32 examples.
    }
    \label{tab:opt_32}
    \vspace{-3pt}
\end{table*}

\setlength{\tabcolsep}{0.2cm}
\begin{table*}[ht]
    \centering
    \resizebox{0.98\textwidth}{!}{
        \begin{tabular}{lcccccccccccccccc}
\toprule
&\multicolumn{3}{c}{\tf{tweet\_eval\_hate}} && \multicolumn{3}{c}{\tf{tweet\_eval\_atheism}} && \multicolumn{3}{c}{\tf{tweet\_eval\_feminist}} && \multicolumn{3}{c}{\tf{sick}} \\
&Random & Abstract & Gold && Random & Abstract & Gold && Random & Abstract & Gold && Random & Abstract & Gold \\
\midrule\
\texttt{7B} & 0.59 & 0.53 & 0.64&& 0.33 & 0.31 & 0.37&& 0.41 & 0.43 & 0.45&& 0.32 & 0.36 & 0.38 \\
\texttt{13B} & 0.63 & 0.53 & 0.65&& 0.31 & 0.34 & 0.28&& 0.43 & 0.34 & 0.44&& 0.39 & 0.41 & 0.41 \\
\texttt{30B} & 0.64 & 0.58 & 0.72&& 0.38 & 0.47 & 0.52&& 0.57 & 0.49 & 0.65&& 0.37 & 0.43 & 0.41 \\
\texttt{65B} & 0.69 & 0.58 & 0.72&& 0.4 & 0.42 & 0.58&& 0.54 & 0.42 & 0.58&& 0.38 & 0.46 & 0.41 \\
\toprule
&\multicolumn{3}{c}{\tf{financial\_phrasebank}} && \multicolumn{3}{c}{\tf{ethos\_race}} && \multicolumn{3}{c}{\tf{ethos\_gender}} && \multicolumn{3}{c}{\tf{ethos\_religion}} \\
&Random & Abstract & Gold && Random & Abstract & Gold && Random & Abstract & Gold && Random & Abstract & Gold \\
\midrule\
\texttt{7B} & 0.31 & 0.37 & 0.41&& 0.43 & 0.45 & 0.32&& 0.36 & 0.38 & 0.64&& 0.4 & 0.7 & 0.65 \\
\texttt{13B} & 0.34 & 0.28 & 0.43&& 0.34 & 0.44 & 0.39&& 0.41 & 0.41 & 0.42&& 0.35 & 0.61 & 0.61 \\
\texttt{30B} & 0.47 & 0.52 & 0.57&& 0.49 & 0.65 & 0.37&& 0.43 & 0.41 & 0.65&& 0.38 & 0.79 & 0.69 \\
\texttt{65B} & 0.42 & 0.58 & 0.54&& 0.42 & 0.58 & 0.38&& 0.46 & 0.41 & 0.6&& 0.44 & 0.83 & 0.69 \\
\toprule
&\multicolumn{3}{c}{\tf{ethos\_national\_origin}} && \multicolumn{3}{c}{\tf{snli}} && \multicolumn{3}{c}{\tf{sst2}} && \multicolumn{3}{c}{\tf{trec}} \\
&Random & Abstract & Gold && Random & Abstract & Gold && Random & Abstract & Gold && Random & Abstract & Gold \\
\midrule\
\texttt{7B} & 0.45 & 0.32 & 0.36&& 0.38 & 0.64 & 0.4&& 0.7 & 0.65 & 0.56&& 0.73 & 0.61 & 0.53 \\
\texttt{13B} & 0.44 & 0.39 & 0.41&& 0.41 & 0.42 & 0.35&& 0.61 & 0.61 & 0.52&& 0.66 & 0.59 & 0.5 \\
\texttt{30B} & 0.65 & 0.37 & 0.43&& 0.41 & 0.65 & 0.38&& 0.79 & 0.69 & 0.52&& 0.76 & 0.65 & 0.52 \\
\texttt{65B} & 0.58 & 0.38 & 0.46&& 0.41 & 0.6 & 0.44&& 0.83 & 0.69 & 0.55&& 0.75 & 0.65 & 0.56 \\
\toprule
&\multicolumn{3}{c}{\tf{rte}} && \multicolumn{3}{c}{\tf{wnli}} && \multicolumn{3}{c}{\tf{mrpc}} && \multicolumn{3}{c}{\tf{poem}} \\
&Random & Abstract & Gold && Random & Abstract & Gold && Random & Abstract & Gold && Random & Abstract & Gold \\
\midrule\
\texttt{7B} & 0.64 & 0.4 & 0.7&& 0.65 & 0.56 & 0.73&& 0.61 & 0.53 & 0.7&& 0.71 & 0.52 & 0.78 \\
\texttt{13B} & 0.42 & 0.35 & 0.61&& 0.61 & 0.52 & 0.66&& 0.59 & 0.5 & 0.64&& 0.71 & 0.54 & 0.78 \\
\texttt{30B} & 0.65 & 0.38 & 0.79&& 0.69 & 0.52 & 0.76&& 0.65 & 0.52 & 0.77&& 0.67 & 0.56 & 0.86 \\
\texttt{65B} & 0.6 & 0.44 & 0.83&& 0.69 & 0.55 & 0.75&& 0.65 & 0.56 & 0.77&& 0.73 & 0.6 & 0.87 \\

\bottomrule

        \end{tabular}
    }
    \caption{
    Single dataset accuracies across the LLaMA model family, using 8 examples.
    }
    \label{tab:llama_8}
    \vspace{-3pt}
\end{table*}

\setlength{\tabcolsep}{0.2cm}
\begin{table*}[ht]
    \centering
    \resizebox{0.98\textwidth}{!}{
        \begin{tabular}{lcccccccccccccccc}
\toprule
&\multicolumn{3}{c}{\tf{tweet\_eval\_hate}} && \multicolumn{3}{c}{\tf{tweet\_eval\_atheism}} && \multicolumn{3}{c}{\tf{tweet\_eval\_feminist}} && \multicolumn{3}{c}{\tf{sick}} \\
&Random & Abstract & Gold && Random & Abstract & Gold && Random & Abstract & Gold && Random & Abstract & Gold \\
\midrule\
\texttt{7B} & 0.61 & 0.58 & 0.66&& 0.33 & 0.49 & 0.37&& 0.41 & 0.35 & 0.45&& 0.31 & 0.43 & 0.36 \\
\texttt{13B} & 0.6 & 0.58 & 0.66&& 0.27 & 0.5 & 0.34&& 0.4 & 0.34 & 0.42&& 0.37 & 0.42 & 0.41 \\
\texttt{30B} & 0.67 & 0.67 & 0.74&& 0.37 & 0.54 & 0.53&& 0.47 & 0.5 & 0.62&& 0.36 & 0.51 & 0.42 \\
\texttt{65B} & 0.66 & 0.62 & 0.73&& 0.37 & 0.56 & 0.6&& 0.52 & 0.53 & 0.6&& 0.38 & 0.55 & 0.42 \\
\toprule
&\multicolumn{3}{c}{\tf{financial\_phrasebank}} && \multicolumn{3}{c}{\tf{ethos\_race}} && \multicolumn{3}{c}{\tf{ethos\_gender}} && \multicolumn{3}{c}{\tf{ethos\_religion}} \\
&Random & Abstract & Gold && Random & Abstract & Gold && Random & Abstract & Gold && Random & Abstract & Gold \\
\midrule\
\texttt{7B} & 0.49 & 0.37 & 0.41&& 0.35 & 0.45 & 0.31&& 0.43 & 0.36 & 0.65&& 0.46 & 0.72 & 0.6 \\
\texttt{13B} & 0.5 & 0.34 & 0.4&& 0.34 & 0.42 & 0.37&& 0.42 & 0.41 & 0.41&& 0.39 & 0.59 & 0.56 \\
\texttt{30B} & 0.54 & 0.53 & 0.47&& 0.5 & 0.62 & 0.36&& 0.51 & 0.42 & 0.64&& 0.49 & 0.84 & 0.6 \\
\texttt{65B} & 0.56 & 0.6 & 0.52&& 0.53 & 0.6 & 0.38&& 0.55 & 0.42 & 0.56&& 0.54 & 0.87 & 0.62 \\
\toprule
&\multicolumn{3}{c}{\tf{ethos\_national\_origin}} && \multicolumn{3}{c}{\tf{snli}} && \multicolumn{3}{c}{\tf{sst2}} && \multicolumn{3}{c}{\tf{trec}} \\
&Random & Abstract & Gold && Random & Abstract & Gold && Random & Abstract & Gold && Random & Abstract & Gold \\
\midrule\
\texttt{7B} & 0.45 & 0.31 & 0.43&& 0.36 & 0.65 & 0.46&& 0.72 & 0.6 & 0.53&& 0.72 & 0.57 & 0.59 \\
\texttt{13B} & 0.42 & 0.37 & 0.42&& 0.41 & 0.41 & 0.39&& 0.59 & 0.56 & 0.51&& 0.66 & 0.59 & 0.5 \\
\texttt{30B} & 0.62 & 0.36 & 0.51&& 0.42 & 0.64 & 0.49&& 0.84 & 0.6 & 0.58&& 0.74 & 0.6 & 0.65 \\
\texttt{65B} & 0.6 & 0.38 & 0.55&& 0.42 & 0.56 & 0.54&& 0.87 & 0.62 & 0.58&& 0.75 & 0.66 & 0.65 \\
\toprule
&\multicolumn{3}{c}{\tf{rte}} && \multicolumn{3}{c}{\tf{wnli}} && \multicolumn{3}{c}{\tf{mrpc}} && \multicolumn{3}{c}{\tf{poem}} \\
&Random & Abstract & Gold && Random & Abstract & Gold && Random & Abstract & Gold && Random & Abstract & Gold \\
\midrule\
\texttt{7B} & 0.65 & 0.46 & 0.72&& 0.6 & 0.53 & 0.72&& 0.57 & 0.59 & 0.67&& 0.65 & 0.59 & 0.78 \\
\texttt{13B} & 0.41 & 0.39 & 0.59&& 0.56 & 0.51 & 0.66&& 0.59 & 0.5 & 0.73&& 0.69 & 0.54 & 0.78 \\
\texttt{30B} & 0.64 & 0.49 & 0.84&& 0.6 & 0.58 & 0.74&& 0.6 & 0.65 & 0.74&& 0.65 & 0.64 & 0.85 \\
\texttt{65B} & 0.56 & 0.54 & 0.87&& 0.62 & 0.58 & 0.75&& 0.66 & 0.65 & 0.78&& 0.73 & 0.64 & 0.85 \\

\bottomrule

        \end{tabular}
    }
    \caption{
    Single dataset accuracies across the LLaMA model family, using 16 examples.
    }
    \label{tab:llama_16}
    \vspace{-3pt}
\end{table*}

\setlength{\tabcolsep}{0.2cm}
\begin{table*}[ht]
    \centering
    \resizebox{0.98\textwidth}{!}{
        \begin{tabular}{lcccccccccccccccc}
\toprule
&\multicolumn{3}{c}{\tf{tweet\_eval\_hate}} && \multicolumn{3}{c}{\tf{tweet\_eval\_atheism}} && \multicolumn{3}{c}{\tf{tweet\_eval\_feminist}} && \multicolumn{3}{c}{\tf{sick}} \\
&Random & Abstract & Gold && Random & Abstract & Gold && Random & Abstract & Gold && Random & Abstract & Gold \\
\midrule\
\texttt{7B} & 0.58 & 0.58 & 0.64&& 0.33 & 0.51 & 0.35&& 0.4 & 0.38 & 0.47&& 0.36 & 0.46 & 0.4 \\
\texttt{13B} & 0.6 & 0.59 & 0.68&& 0.3 & 0.46 & 0.37&& 0.41 & 0.42 & 0.46&& 0.36 & 0.42 & 0.42 \\
\texttt{30B} & 0.65 & 0.64 & 0.73&& 0.32 & 0.53 & 0.6&& 0.48 & 0.51 & 0.63&& 0.35 & 0.55 & 0.42 \\
\texttt{65B} & 0.64 & 0.68 & 0.78&& 0.38 & 0.51 & 0.6&& 0.45 & 0.49 & 0.63&& 0.36 & 0.62 & 0.43 \\
\toprule
&\multicolumn{3}{c}{\tf{financial\_phrasebank}} && \multicolumn{3}{c}{\tf{ethos\_race}} && \multicolumn{3}{c}{\tf{ethos\_gender}} && \multicolumn{3}{c}{\tf{ethos\_religion}} \\
&Random & Abstract & Gold && Random & Abstract & Gold && Random & Abstract & Gold && Random & Abstract & Gold \\
\midrule\
\texttt{7B} & 0.51 & 0.35 & 0.4&& 0.38 & 0.47 & 0.36&& 0.46 & 0.4 & 0.64&& 0.5 & 0.74 & 0.61 \\
\texttt{13B} & 0.46 & 0.37 & 0.41&& 0.42 & 0.46 & 0.36&& 0.42 & 0.42 & 0.38&& 0.38 & 0.56 & 0.65 \\
\texttt{30B} & 0.53 & 0.6 & 0.48&& 0.51 & 0.63 & 0.35&& 0.55 & 0.42 & 0.61&& 0.61 & 0.88 & 0.66 \\
\texttt{65B} & 0.51 & 0.6 & 0.45&& 0.49 & 0.63 & 0.36&& 0.62 & 0.43 & 0.52&& 0.66 & 0.88 & 0.59 \\
\toprule
&\multicolumn{3}{c}{\tf{ethos\_national\_origin}} && \multicolumn{3}{c}{\tf{snli}} && \multicolumn{3}{c}{\tf{sst2}} && \multicolumn{3}{c}{\tf{trec}} \\
&Random & Abstract & Gold && Random & Abstract & Gold && Random & Abstract & Gold && Random & Abstract & Gold \\
\midrule\
\texttt{7B} & 0.47 & 0.36 & 0.46&& 0.4 & 0.64 & 0.5&& 0.74 & 0.61 & 0.59&& 0.67 & 0.47 & 0.62 \\
\texttt{13B} & 0.46 & 0.36 & 0.42&& 0.42 & 0.38 & 0.38&& 0.56 & 0.65 & 0.53&& 0.73 & 0.67 & 0.57 \\
\texttt{30B} & 0.63 & 0.35 & 0.55&& 0.42 & 0.61 & 0.61&& 0.88 & 0.66 & 0.6&& 0.74 & 0.55 & 0.6 \\
\texttt{65B} & 0.63 & 0.36 & 0.62&& 0.43 & 0.52 & 0.66&& 0.88 & 0.59 & 0.63&& 0.76 & 0.58 & 0.66 \\
\toprule
&\multicolumn{3}{c}{\tf{rte}} && \multicolumn{3}{c}{\tf{wnli}} && \multicolumn{3}{c}{\tf{mrpc}} && \multicolumn{3}{c}{\tf{poem}} \\
&Random & Abstract & Gold && Random & Abstract & Gold && Random & Abstract & Gold && Random & Abstract & Gold \\
\midrule\
\texttt{7B} & 0.64 & 0.5 & 0.74&& 0.61 & 0.59 & 0.67&& 0.47 & 0.62 & 0.69&& 0.65 & 0.64 & 0.79 \\
\texttt{13B} & 0.38 & 0.38 & 0.56&& 0.65 & 0.53 & 0.73&& 0.67 & 0.57 & 0.76&& 0.7 & 0.62 & 0.83 \\
\texttt{30B} & 0.61 & 0.61 & 0.88&& 0.66 & 0.6 & 0.74&& 0.55 & 0.6 & 0.8&& 0.57 & 0.65 & 0.82 \\
\texttt{65B} & 0.52 & 0.66 & 0.88&& 0.59 & 0.63 & 0.76&& 0.58 & 0.66 & 0.77&& 0.63 & 0.73 & 0.87 \\
\bottomrule

        \end{tabular}
    }
    \caption{
    Single dataset accuracies across the LLaMA model family, using 32 examples.
    }
    \label{tab:llama_32}
    \vspace{-3pt}
\end{table*}

\end{document}